%% file: main.tex
\documentclass[10pt]{article} 
\usepackage[preprint]{tmlr}

\input{math_commands.tex}

\usepackage[pagebackref,breaklinks,colorlinks,linkcolor={blue},citecolor={blue}]{hyperref}

\usepackage{url}

\usepackage{booktabs}
\usepackage{graphicx}
\usepackage{multirow}
\usepackage{xcolor}
\usepackage{tikz}
\usetikzlibrary{positioning, arrows.meta, calc, fit, backgrounds}

\usepackage{enumitem}
\setlist{topsep=5pt,itemsep=0pt}

\newcommand{\chimera}{\textsc{Chimera-Bench}}

\title{\chimera{}: A Benchmark Dataset for Epitope-Specific Antibody Design}

\author{
  Mansoor Ahmed~\textsuperscript{1,2\thanks{Address correspondence to: \quad \texttt{mahmed76@student.gsu.edu}, \quad \texttt{mpatterson30@gsu.edu} }}  ,
  Nadeem Taj~\textsuperscript{3},
  Imdad Ullah Khan\textsuperscript{4},
  Hemanth Venkateswara\textsuperscript{1}, \\[6pt]
  \hfill \textbf{Murray Patterson}\textsuperscript{1\tiny{*}} \hfill \\[8pt]
  \normalfont\normalsize
  \textsuperscript{1}\textit{Georgia State University, Atlanta, GA, USA} \\
  \textsuperscript{2}\textit{Georgia Institute of Technology, Atlanta, GA, USA} \\
  \textsuperscript{3}\textit{University of Engineering and Technology, Lahore, Pakistan} \\
  \textsuperscript{4}\textit{Lahore University of Management Sciences, Pakistan}
}

\def\month{MM}  
\def\year{YYYY} 
\def\openreview{\url{https://openreview.net/forum?id=XXXX}} 

\begin{document}

\maketitle

\begin{abstract}
Computational antibody design has seen rapid methodological progress, with dozens of deep generative methods proposed in the past three years, yet the field lacks a standardized benchmark for fair comparison and model development.
These methods are evaluated on different SAbDab snapshots, non-overlapping test sets, and incompatible metrics, and the literature fragments the design problem into numerous sub-tasks with no common definition.
We introduce \chimera{} (\textbf{C}DR \textbf{M}odeling with \textbf{E}pitope-guided \textbf{R}edesign), a unified benchmark built around a single canonical task: \emph{epitope-conditioned CDR sequence--structure co-design}.
\chimera{} provides three components. The first is a curated, deduplicated dataset of \textbf{2,922} antibody--antigen complexes with epitope and paratope annotations. The second is a set of three biologically motivated splits that test generalization to unseen epitopes, unseen antigen folds, and prospective temporal targets. The third is a comprehensive evaluation protocol with five metric groups, including novel epitope-specificity measures.
We benchmark eleven methods spanning six generative paradigms and report results across all splits.
\chimera{} is the largest dataset of its kind for the antibody design problem, allowing the community to develop and test novel methods and evaluate their generalizability. The source code and data are available at: \url{https://github.com/mansoor181/chimera-bench.git}
\end{abstract}

\section{Introduction}
\label{sec:intro}

Antibodies are among the most important classes of biotherapeutics~\citep{norman2020computational}. Their binding specificity is largely determined by six complementarity-determining regions (CDRs), particularly CDR-H3, which makes these loops the primary targets for computational design. Deep generative models have recently transformed this space, with diffusion models~\citep{luo2022diffab,martinkus2023abdiffuser}, flow matching~\citep{tan2025dyab}, equivariant graph neural networks~\citep{wu2025raad,kong2022mean,kong2023dymean}, autoregressive models~\citep{jin2021refinegnn}, and foundation models~\citep{wang2025iggm} all proposed for CDR sequence--structure co-design. Yet despite this rapid progress, there is no standardized way to compare these methods. They are trained on different snapshots of the Structural Antibody Database (SAbDab)~\citep{dunbar2014sabdab} with varying filtering criteria, evaluated on non-overlapping test sets from the RAbD dataset~\citep{adolf2018rabd} to hold-outs, and scored with incompatible metrics computed under different definitions. For example, contact cutoffs vary from 4.5 to 6.6~\AA{} across methods, and RMSD is computed with or without Kabsch alignment. The methods also require inputs in different and incompatible formats, making head-to-head comparison impractical without building separate data pipelines for each method.

A more fundamental issue is the absence of a common task definition. The literature fragments antibody design into numerous sub-tasks, including inverse folding, structure prediction, co-design, docking, affinity optimization, de novo generation, and epitope-conditioned design, and individual methods address different subsets. This makes it unclear what is being compared, even when two papers report the same metric name. We argue that in many therapeutic settings, the target epitope is specified and propose \emph{epitope-conditioned CDR sequence--structure co-design} as the canonical formulation. Given an antigen structure, an epitope specification, and an antibody framework, the task is to design CDR residues that are structurally valid, contact the target epitope, and avoid off-target binding. This formulation subsumes the sub-tasks above as special cases and accommodates all baseline methods. Methods that natively support epitope conditioning map directly to this definition, while the remaining methods can be evaluated as-is, with epitope-specificity metrics revealing whether their designs contact the intended site.

We introduce \chimera{} (\textbf{C}DR \textbf{M}odeling with \textbf{E}pitope-guided \textbf{R}edesign), a benchmark dataset for the antibody design task. Our contributions are:
\begin{itemize}
    \item A \textbf{canonical problem definition} that unifies seven sub-tasks from the literature into a single epitope-conditioned CDR co-design formulation.
    \item A \textbf{curated dataset} of deduplicated, quality-filtered antibody--antigen complexes from SAbDab, CDR masks, epitope/paratope annotations, and contact maps.
    \item \textbf{Three evaluation splits} (epitope-group, antigen-fold, and temporal), each with cluster-level leakage prevention, testing generalization along distinct biological axes.
    \item A \textbf{comprehensive evaluation protocol} with five metric groups covering sequence quality, structural accuracy, binding interface quality, epitope specificity, and designability, including novel epitope-specificity metrics not reported by any existing method.
    \item The \textbf{first head-to-head retraining} of eleven methods spanning six generative paradigms under identical data, splits, and metrics, together with the empirical findings this comparison exposes.
\end{itemize}


\section{Related Work}
\label{sec:related}

\paragraph{Antibody CDR design methods.}
Early learning-based approaches to CDR design were autoregressive. RefineGNN~\citep{jin2021refinegnn} generates CDR residues left-to-right while iteratively refining the predicted structure, and AbDockGen~\citep{jin2022hern} combines hierarchical equivariant refinement with antibody--antigen docking. MEAN~\citep{kong2022mean} recast the problem as E(3)-equivariant graph translation, and dyMEAN~\citep{kong2023dymean} extended this to end-to-end full-atom antibody design. RAAD~\citep{wu2025raad} later introduced relation-aware equivariance for settings where the epitope is unknown. Diffusion models have since become the dominant paradigm, beginning with DiffAb~\citep{luo2022diffab}, which pioneered antigen-conditioned CDR diffusion over sequences, coordinates, and orientations simultaneously. AbX~\citep{zhu2024abx} augmented this framework with evolutionary and physical constraints, and LEAD~\citep{yao2025lead} introduced property-guided sampling through black-box optimization in a shared latent space. dyAb~\citep{tan2025dyab} applies flow matching to flexible antibody design using AlphaFold-predicted pre-binding antigen conformations, and AbODE~\citep{verma2023abode} formulates CDR generation through conjoined ODEs. 
RFAntibody~\citep{bennett2025rfdiffusion-ab} fine-tunes RFdiffusion on ${\sim}$8{,}100 PDB antibody structures for de novo design of complete variable regions (VHHs and scFvs) conditioned on a target epitope. In the broader protein design space, BoltzGen~\citep{stark2025boltzgen} is an all-atom diffusion model that unifies binder design and structure prediction across modalities, including nanobodies, though it targets general binder design rather than CDR-level co-design. ProteinMPNN~\citep{dauparas2022proteinmpnn}, while also not antibody-specific, serves as a strong inverse folding baseline for CDR sequence design given a fixed backbone.

\paragraph{Existing benchmarks and datasets.}
SAbDab~\citep{dunbar2014sabdab} is the primary structural database for antibodies but provides no standardized splits, evaluation protocol, or task definition for generative design and contains redundant complexes that require preprocessing for training neural networks. The RAbD benchmark~\citep{adolf2018rabd} defines 60 antibody--antigen complexes for CDR-H3 redesign but lacks epitope annotations and modern generative metrics such as DockQ or diversity. SKEMPI v2~\citep{jankauskaite2019skempi} provides experimental binding affinity measurements for protein complexes including approximately 53 antibody entries, but targets mutation-level affinity prediction rather than generative design. AbBiBench~\citep{zhao2025abbibench} benchmarks antibody binding affinity maturation with 184{,}500+ experimental measurements across 14 antibodies, but focuses on affinity prediction rather than generative CDR design. In the broader protein domain, ProteinGym~\citep{notin2023proteingym} benchmarks fitness prediction across 250+ deep mutational scanning assays and ATOM3D~\citep{townshend2020atom3d} defines molecular learning tasks on 3D structures, but neither addresses the antibody-specific challenges of CDR co-design, epitope conditioning, or binding interface evaluation. To our knowledge, \chimera{} is the first benchmark to combine standardized data curation, biologically motivated splits, epitope-specificity metrics, and cross-method format compatibility for computational antibody design.

Table~\ref{tab:benchmark_comparison} positions \chimera{} against prior antibody and protein design benchmarks (RAbD~\citep{adolf2018rabd}, SAbDab~\citep{dunbar2014sabdab}, SKEMPI v2~\citep{jankauskaite2019skempi}, AsEP~\citep{liu2024asep}, AbBiBench~\citep{zhao2025abbibench}, ProteinGym~\citep{notin2023proteingym}). \chimera{} is the largest curated benchmark of its kind and the first to combine standardized annotations, multiple evaluation splits, and cross-method format compatibility.

\begin{table}[h!]
\centering
\caption{Comparison of \chimera{} with prior antibody design benchmarks and datasets.}
\label{tab:benchmark_comparison}
\scriptsize
\setlength{\tabcolsep}{3pt}
\begin{tabular}{lccccccc}
\toprule
 & \textbf{\chimera{}} & RAbD & SAbDab & SKEMPI v2 & AsEP & AbBiBench & ProteinGym \\
\midrule
Domain & Ab design & Ab design & Ab structures & Affinity & Epitope pred. & Affinity mat. & Fitness \\
Complexes & 2,922 & 60 & 7{,}000+ & 53 (Ab) & 1{,}723 & 14 Ab / 9 Ag & -- \\
Epitope labels & \checkmark & -- & -- & -- & \checkmark & -- & -- \\
Paratope labels & \checkmark & -- & -- & -- & \checkmark & -- & -- \\
CDR annotations & \checkmark & -- & -- & -- & -- & -- & -- \\
Multiple splits & 3 & 1 & -- & 1 & 2 & 3 & 3 \\
Leakage prevention & \checkmark & -- & -- & -- & \checkmark & \checkmark & \checkmark \\
Affinity data & -- & -- & Partial & \checkmark & -- & \checkmark & \checkmark \\
\bottomrule
\end{tabular}
\end{table}

\section{The \chimera{} Dataset}
\label{sec:benchmark}

\subsection{Dataset Construction}
\label{sec:dataset}

The \chimera{} dataset is constructed from SAbDab~\citep{dunbar2014sabdab} through an eight-step pipeline (Figure~\ref{fig:pipeline}). Starting from 20{,}509 SAbDab entries, we apply quality filters (protein/peptide antigens only, paired VH/VL chains, resolution $\leq$4.0~\AA), cluster at 95\% sequence identity using MMseqs2~\citep{steinegger2017mmseqs2}, and validate each complex for ANARCI~\citep{dunbar2016anarci} numberability, conserved residue presence, CDR completeness, and backbone integrity. Residues are numbered under both the IMGT~\citep{lefranc2003imgt} and Chothia~\citep{chothia1987canonical} schemes, and epitope/paratope residues are identified at a 4.5~\AA{} contact cutoff. The full processing details, validation criteria, and exclusion breakdown are provided in Appendix~\ref{app:processing}.

The final dataset contains \textbf{2,922 complexes} from 2,721 unique PDBs (2,485 protein-antigen, 437 peptide-antigen, median resolution 2.72~\AA), with on average 17.9 epitope and 20.5 paratope residues per complex. The graph construction details are in Appendix~\ref{app:graphs}.

\begin{figure}[t]
\centering
\resizebox{\textwidth}{!}{%
\begin{tikzpicture}[
    node distance=0.35cm and 0.4cm,
    stage/.style={
        rectangle, rounded corners=2.5pt, draw=black!65, thick,
        minimum width=1.9cm, minimum height=0.75cm,
        font=\footnotesize\sffamily, text=black, align=center,
        fill=#1,
    },
    stage/.default=white,
    arr/.style={-{Stealth[length=4pt]}, thick, black!55},
    cntlbl/.style={font=\scriptsize\sffamily\bfseries, text=black!60},
    detlbl/.style={font=\tiny\sffamily, text=black!45, align=center},
    phaselbl/.style={font=\tiny\sffamily\bfseries, text=black!40},
]

\node[stage=blue!6] (collect) {Collect};
\node[cntlbl, below=1pt of collect] {20,509};
\node[detlbl, above=2pt of collect] {SAbDab + PDB};

\node[stage=orange!8, right=0.7cm of collect] (filter) {Filter};
\node[cntlbl, below=1pt of filter] {9,171};
\node[detlbl, above=2pt of filter] {Res.\ $\leq$4.0\,\AA\\paired VH/VL};

\node[stage=orange!8, right=0.7cm of filter] (dedup) {De-duplicate};
\node[cntlbl, below=1pt of dedup] {2,981};
\node[detlbl, above=2pt of dedup] {MMseqs2\\95\% seq.\ id.};

\node[stage=violet!8, right=0.7cm of dedup] (annotate) {Annotate};
\node[detlbl, above=2pt of annotate] {IMGT + Chothia\\CDR masks};

\node[stage=violet!8, right=0.7cm of annotate] (contacts) {Contacts};
\node[detlbl, above=2pt of contacts] {Epitope/paratope\\4.5\,\AA\ cutoff};

\node[stage=red!6, below=1.3cm of contacts] (validate) {Validate};
\node[cntlbl, below=1pt of validate] {2,922};
\node[detlbl, above=2pt of validate] {Numbering\\backbone QC};

\node[stage=cyan!8, left=0.7cm of validate] (features) {Features};
\node[detlbl, above=2pt of features] {Atom14 coords\\contact maps};

\node[stage=cyan!8, below left=0.05cm and 0.8cm of features] (graphs) {Graphs};
\node[detlbl, left=4pt of graphs] {Heterogeneous\\residue graphs};

\node[stage=cyan!8, above left=0.05cm and 0.8cm of features] (splits) {Splits};
\node[detlbl, left=4pt of splits] {Epitope-group\\Antigen-fold\\Temporal};


\draw[arr] (collect) -- (filter);
\draw[arr] (filter) -- (dedup);
\draw[arr] (dedup) -- (annotate);
\draw[arr] (annotate) -- (contacts);
\draw[arr] (contacts) -- (validate);
\draw[arr] (validate) -- (features);

\draw[arr] (features.west) -- ++(0,0.02) -| (splits.south);
\draw[arr] (features.west) -- ++(0,-0.02) -| (graphs.north);

\begin{scope}[on background layer]
    \node[fit=(collect)(filter)(dedup), inner sep=4pt, fill=black!2, rounded corners=3pt] {};
    \node[fit=(annotate)(contacts)(validate), inner sep=4pt, fill=black!2, rounded corners=3pt] {};
\end{scope}

\end{tikzpicture}%
}
\caption{The \chimera{} data curation pipeline that collects antigen-antibody complexes from SAbDab and produces filtered and annotated complexes for the antibody design task.}
\label{fig:pipeline}
\end{figure}

\subsection{Dataset Statistics}
\label{sec:stats}

\paragraph{Structural properties.} The final dataset comprises 2,922 complexes drawn from 2,721 unique PDB entries at a median resolution of 2.72~\AA. Of these, 2,485 (85\%) target protein antigens and 437 (15\%) target peptide antigens, and the set is dominated by X-ray diffraction (1{,}929) and cryo-EM (989) structures. Antigen lengths vary widely (288$\pm$278 residues), with longer antigens corresponding to viral surface proteins and shorter ones to peptide epitopes. Antibody chains are far more uniform: the IMGT-numbered variable domains span 121$\pm$5 (VH) and 108$\pm$3 (VL) residues, within full heavy and light chains.

\paragraph{Interface properties.} Epitope sizes range from 1 to over 50 residues (mean 17.9), and paratope sizes are slightly larger (mean 20.5), consistent with the observation that paratope residues span multiple CDR loops while epitope residues cluster on a single antigen surface patch. The mean of 45.6 residue-level contact pairs per complex provides a rich signal for evaluating binding interface reconstruction.

\begin{figure}[h!]
\centering
\includegraphics[width=\textwidth]{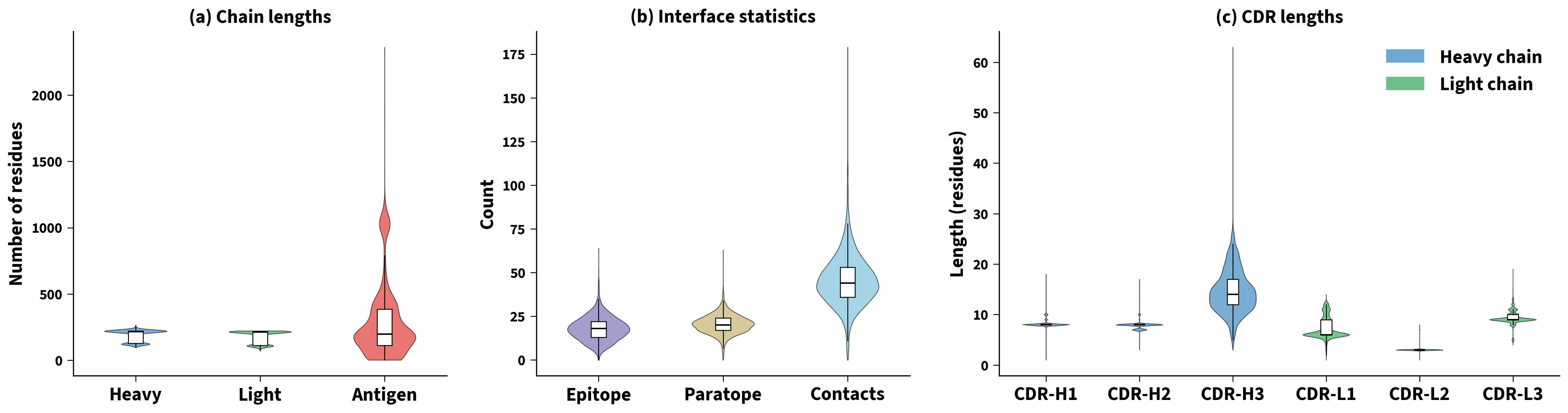}
\caption{Distributions across the \chimera{} dataset. (a) Heavy, light, and antigen chain lengths. (b) Interface statistics, namely epitope size, paratope size, and number of residue contacts per complex. (c) CDR loop lengths under IMGT numbering, where CDR-H3 shows the widest variability consistent with its role as the primary determinant of antigen specificity.}
\label{fig:dataset_overview}
\end{figure}

\paragraph{CDR statistics.} Under IMGT numbering, CDR-H3 is the most variable loop in both length and sequence (mean 14.6$\pm$4.3 residues, range 3--63), with a heavy right tail reflecting the diversity of VDJ recombination. CDR-H1 (8.2$\pm$0.7) and CDR-H2 (7.9$\pm$0.7) show narrower distributions consistent with their more conserved canonical structures. Light chain CDRs are generally more uniform in length: L1 (7.4$\pm$2.1), L2 (3.0$\pm$0.4, nearly fixed-length under IMGT), and L3 (9.4$\pm$1.3).
In our dataset, light-chain residues account for a mean of 35\% of paratope contacts (IQR 25--46\%), consistent with prior analyses~\citep{ramaraj2012antigen}, and are important for overall interface quality.
The full CDR length distributions, together with chain-length and interface-size distributions, are shown in Figure~\ref{fig:dataset_overview}.

\begin{figure}[h!]
\centering
\includegraphics[width=\textwidth]{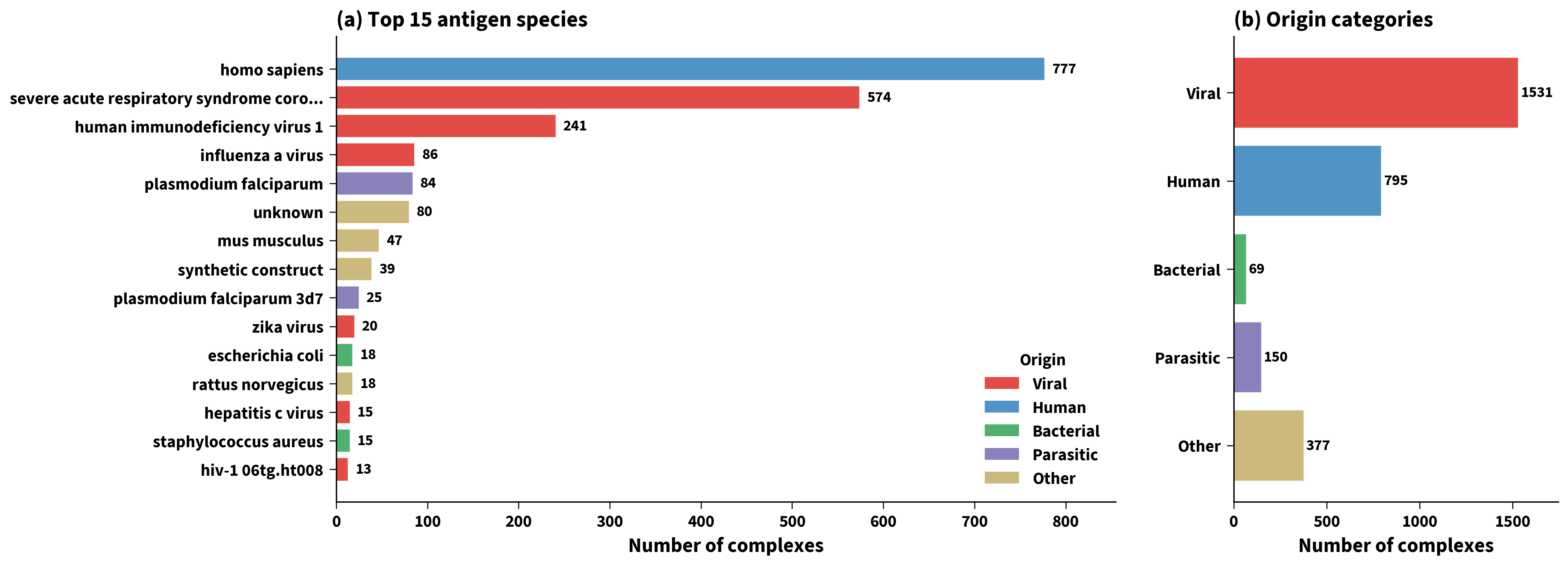}
\caption{Antigen species and origin distributions. (a) Top 15 species by complex count. (b) Breakdown by biological origin category.}
\label{fig:antigen_species}
\end{figure}

\paragraph{Temporal and species distributions.} PDB deposition dates range from the mid-1990s to mid-2024, with a strong concentration in 2019--2023 reflecting the surge in COVID-related antibody structures. The majority of antigens originate from viral proteins or human self-antigens (Figure~\ref{fig:antigen_species}).

\subsection{Epitope Diversity}
\label{sec:diversity}

\begin{figure}[h!]
\centering
\includegraphics[width=\textwidth]{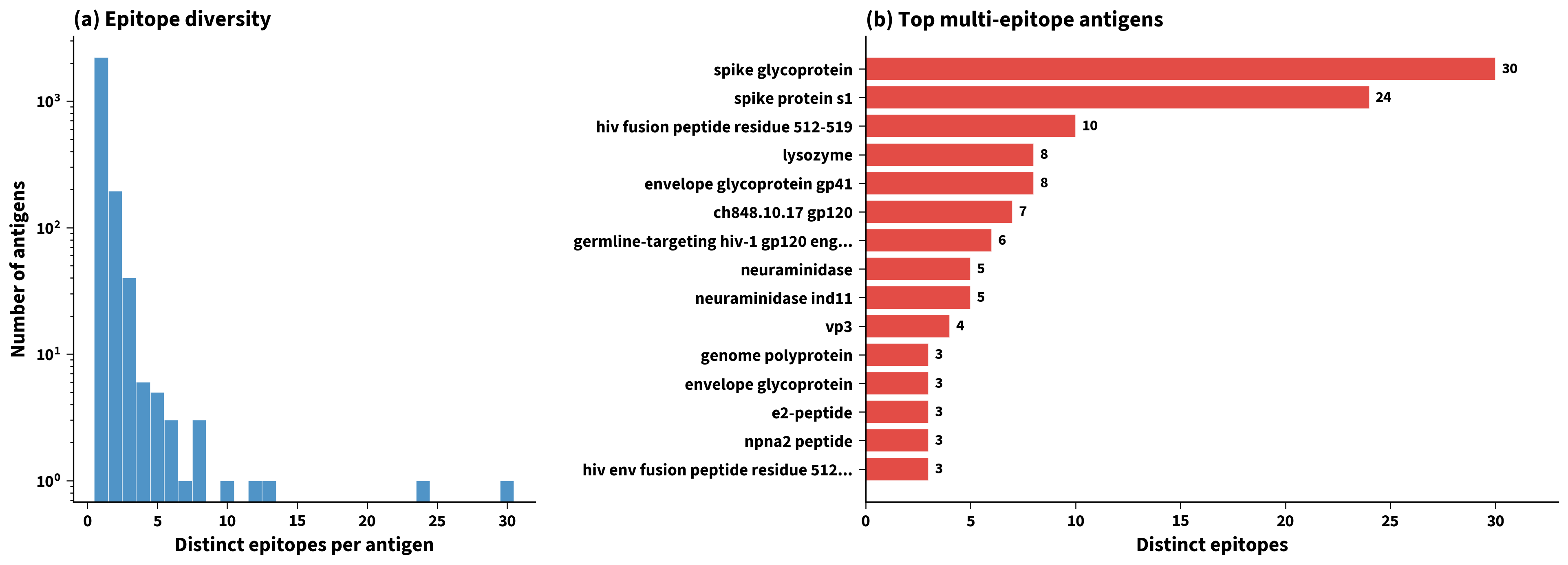}
\caption{Epitope diversity in \chimera{}. (a) Distribution of distinct epitopes per antigen sequence (log scale); most antigens have a single epitope, but a long tail extends to 30. (b) Top 15 multi-epitope antigens ranked by number of distinct epitope sites.}
\label{fig:multi_epitope}
\end{figure}

A distinctive feature of \chimera{} is that many antigens are targeted by multiple antibodies binding to distinct epitope regions. Of the 2{,}470 unique antigen sequences in the dataset, 258 (10.4\%) are bound at two or more distinct epitopes, collectively accounting for 710 of the 2{,}922 complexes (24.3\%). The most promiscuous antigen is targeted at 30 distinct epitope sites (Figure~\ref{fig:multi_epitope}). This epitope diversity is central to \chimera's evaluation. A successful model must not merely reconstruct a plausible binding interface but must do so for the \emph{correct} epitope, which makes epitope-conditioned generation essential. The epitope-group split explicitly tests this capability by holding out entire epitope clusters at test time.

\subsection{Data Splits}
\label{sec:splits}

A central design choice in \chimera{} is the use of three biologically motivated splits, each testing a different generalization axis while enforcing cluster-level separation to prevent data leakage, as shown in Table~\ref{tab:split_summary}. 
The \textbf{epitope-group} split clusters complexes that share the same set of epitope residue positions on the antigen surface, so that the test set contains epitope patterns never seen during training. Two complexes belong to the same cluster if and only if their sorted epitope residue identifiers (chain, position) are identical, which naturally handles discontinuous epitopes. The \textbf{antigen-fold} split groups complexes by antigen identity, ensuring the test set contains entirely unseen antigen targets. The \textbf{temporal} split assigns complexes by PDB deposition date, simulating prospective deployment on structures deposited after the training period.

\begin{table}[h!]
\centering
\caption{Summary of \chimera{} evaluation splits. }
\label{tab:split_summary}
\small
\begin{tabular}{lrrrl}
\toprule
Split & Train & Val & Test & Generalization target \\
\midrule
Epitope-group & 2,338 & 292 & 292 & Unseen epitope patterns \\
Antigen-fold  & 2,338 & 292 & 292 & Unseen antigen topologies \\
Temporal      & 2,337 & 292 & 293 & Prospective (post-2023) \\
\bottomrule
\end{tabular}
\end{table}

\begin{figure}[h!]
\centering
\includegraphics[width=\textwidth]{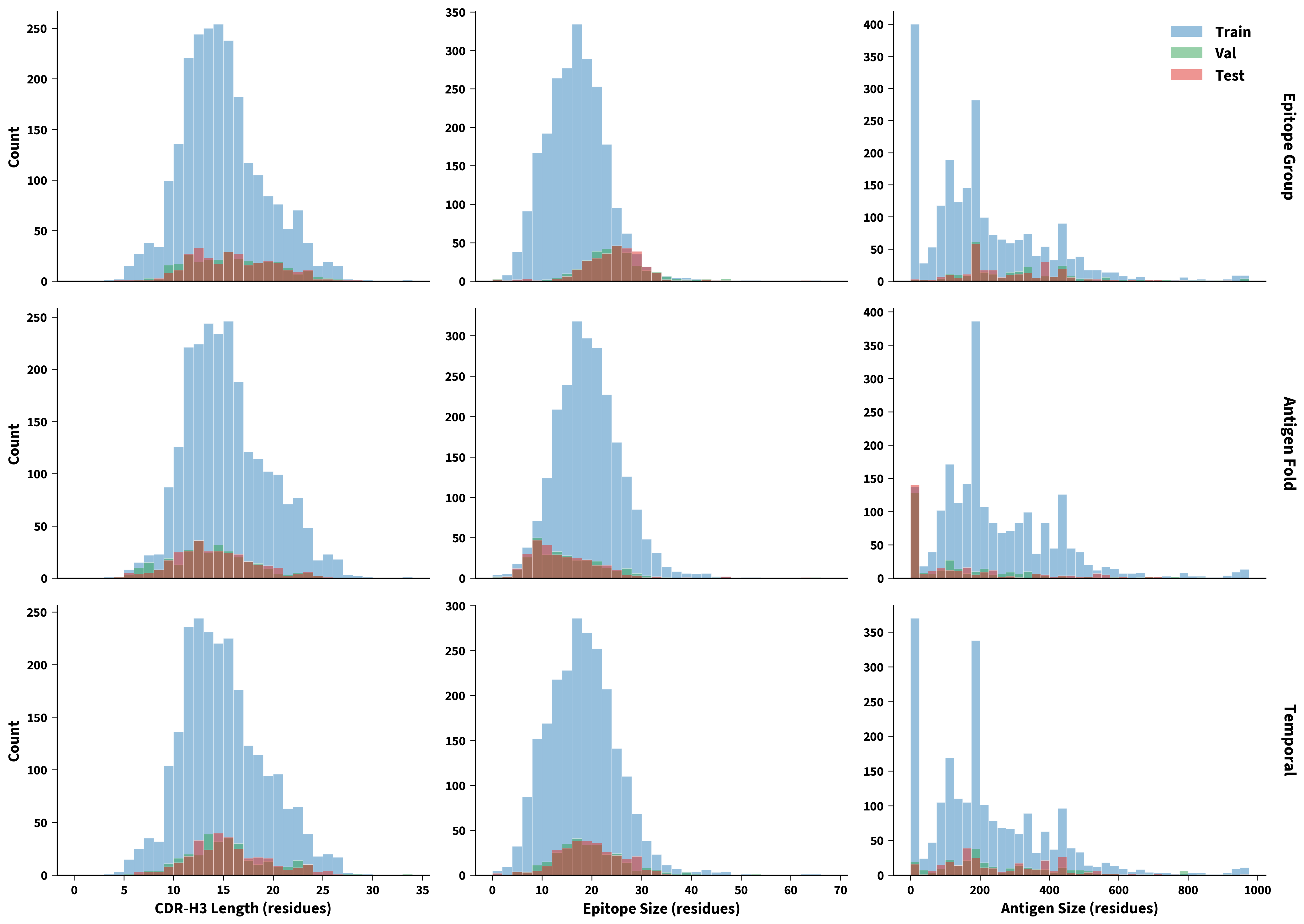}
\caption{Distribution of CDR-H3 length, epitope size, and antigen size (columns) across the train, validation, and test partitions of the three splits (rows). }
\label{fig:split_boxplots}
\end{figure}

The three splits are distribution-matched on the covariates that govern design difficulty, namely CDR-H3 length, epitope size, and antigen size (Figure~\ref{fig:split_boxplots}). This matching ensures that differences in test performance across splits reflect the held-out biological novelty of each axis rather than a covariate shift in the test pool.

\section{Task and Evaluation Protocol}
\label{sec:task_eval}

\subsection{Task Definition}
\label{sec:task}

As discussed above, the antibody design literature fragments the problem into numerous sub-tasks, and individual methods address different subsets. \chimera{} consolidates these into a single canonical task, \emph{epitope-conditioned CDR sequence--structure co-design}, motivated by the observation that in many therapeutic settings the target epitope is specified. Given an antigen structure $A = \{(s_j, \mathbf{x}_j) \mid j \in V_A\}$, an epitope $E \subseteq V_A$, and an antibody framework $F = \{(s_i, \mathbf{x}_i) \mid i \in V_\text{FR}\}$, the task is to design CDR residues that maximize the learned conditional distribution while satisfying epitope contact and precision constraints. Following~\citet{luo2022diffab}, each residue is represented as a tuple of amino acid type, C$\alpha$ coordinate, and local frame orientation. The remaining sub-tasks from the literature are subsumed as special cases: inverse folding fixes the backbone and designs sequence only (the ProteinMPNN setting), structure prediction fixes the sequence, unconditional co-design sets $E = V_A$ (i.e., the entire antigen surface is treated as the target, imposing no epitope constraint), and docking quality is captured implicitly through DockQ and iRMSD.

The task can be expressed concisely as:
\begin{equation}
    R^* = \argmax_{R} \; p_\theta\!\bigl(R \mid A, E, F\bigr), \quad
    \text{s.t.} \;\; \mathcal{C}(R, A) \cap E \neq \emptyset, \;\; \mathcal{C}(R, A) \subseteq E
    \label{eq:task}
\end{equation}
where $R = \{(s_k, \mathbf{x}_k, \mathbf{O}_k)\}_{k \in V_\text{CDR}}$ denotes the designed CDR residues with amino acid type $s_k$, C$\alpha$ coordinate $\mathbf{x}_k$, and local frame orientation $\mathbf{O}_k$, and $\mathcal{C}(R, A) = \{j \in V_A \mid \exists\, k \in V_\text{CDR}: \|\mathbf{x}_k - \mathbf{x}_j\| < d_c\}$ is the set of antigen residues contacted by the designed CDRs within a cutoff distance $d_c$.

\subsection{Evaluation Protocol}
\label{sec:evaluation}

We evaluate designs across five metric groups that collectively assess sequence quality, structural accuracy, binding interface quality, epitope specificity, and designability. 
\textbf{Sequence quality} is measured by amino acid recovery (AAR) and contact AAR (CAAR), which restricts this calculation to paratope residues within 4.5~\AA{} of the antigen, along with perplexity (PPL) for models that produce sequence likelihoods. \textbf{Structural quality} is assessed by C$\alpha$ RMSD after Kabsch alignment and TM-score~\citep{zhang2004tm}. We standardize on Kabsch-aligned RMSD throughout, as existing methods use inconsistent implementations. The \textbf{binding interface quality} is captured by the fraction of native contacts recovered (Fnat), interface RMSD (iRMSD) restricted to contact residues, and the DockQ composite score~\citep{basu2016dockq}, which combines Fnat, iRMSD, and ligand RMSD into a single metric.
\textbf{Epitope specificity} is a novel evaluation axis introduced by \chimera{}. We compute epitope precision, recall, and F1, measuring whether a design contacts the intended binding site. \textbf{Designability} counts known sequence liability motifs associated with manufacturing issues. We use two contact definitions for two distinct purposes. Dataset annotations and the paratope mask use a strict 4.5~\AA{} heavy-atom cutoff, which identifies physical contacts for graph construction and CAAR. The interface and epitope metrics (Fnat, iRMSD, DockQ, EpiF1) instead use an 8~\AA{} C$\alpha$--C$\alpha$ cutoff computed symmetrically on both native and predicted structures, a definition that tolerates the C$\alpha$-level coordinate error of generative models without rewarding spurious atom-level clashes.


We further define two evaluation tracks. 
\begin{enumerate}
    \item \textbf{CDR-H3 co-design} is mandatory for all methods, since CDR-H3 is the most variable loop and the primary determinant of antigen specificity.
    \item \textbf{All-CDR co-design} evaluates methods that support multi-loop generation, designing all six CDRs (H1--H3, L1--L3) simultaneously. 
\end{enumerate}

Few baseline methods natively support epitope conditioning and map directly to the canonical definition.  
Our epitope-specificity metrics then reveal whether their designs contact the intended binding site. 
The formal definitions are provided in Appendix~\ref{app:metrics}.


\section{Experiments}
\label{sec:experiments}

We benchmark eleven antibody design methods spanning six generative paradigms: equivariant GNNs (RAAD~\citep{wu2025raad}, MEAN~\citep{kong2022mean}, dyMEAN~\citep{kong2023dymean}), diffusion models and their extensions (DiffAb~\citep{luo2022diffab}, AbMEGD~\citep{chen2025AbMEGD}, RADAb~\citep{wang2024radab}, AbFlowNet~\citep{abir2025abflownet}), flow matching (dyAb~\citep{tan2025dyab}), autoregressive generation (RefineGNN~\citep{jin2021refinegnn}), hierarchical equivariant networks (AbDockGen~\citep{jin2022hern}), and conjoined ODEs (AbODE~\citep{verma2023abode}). Each method is retrained on \chimera{} using the authors' released code with default hyperparameters. The training details are in Appendix~\ref{app:retraining}.

\subsection{CDR-H3 Design Results}

\begin{figure}[h!]
\centering
\includegraphics[width=\textwidth]{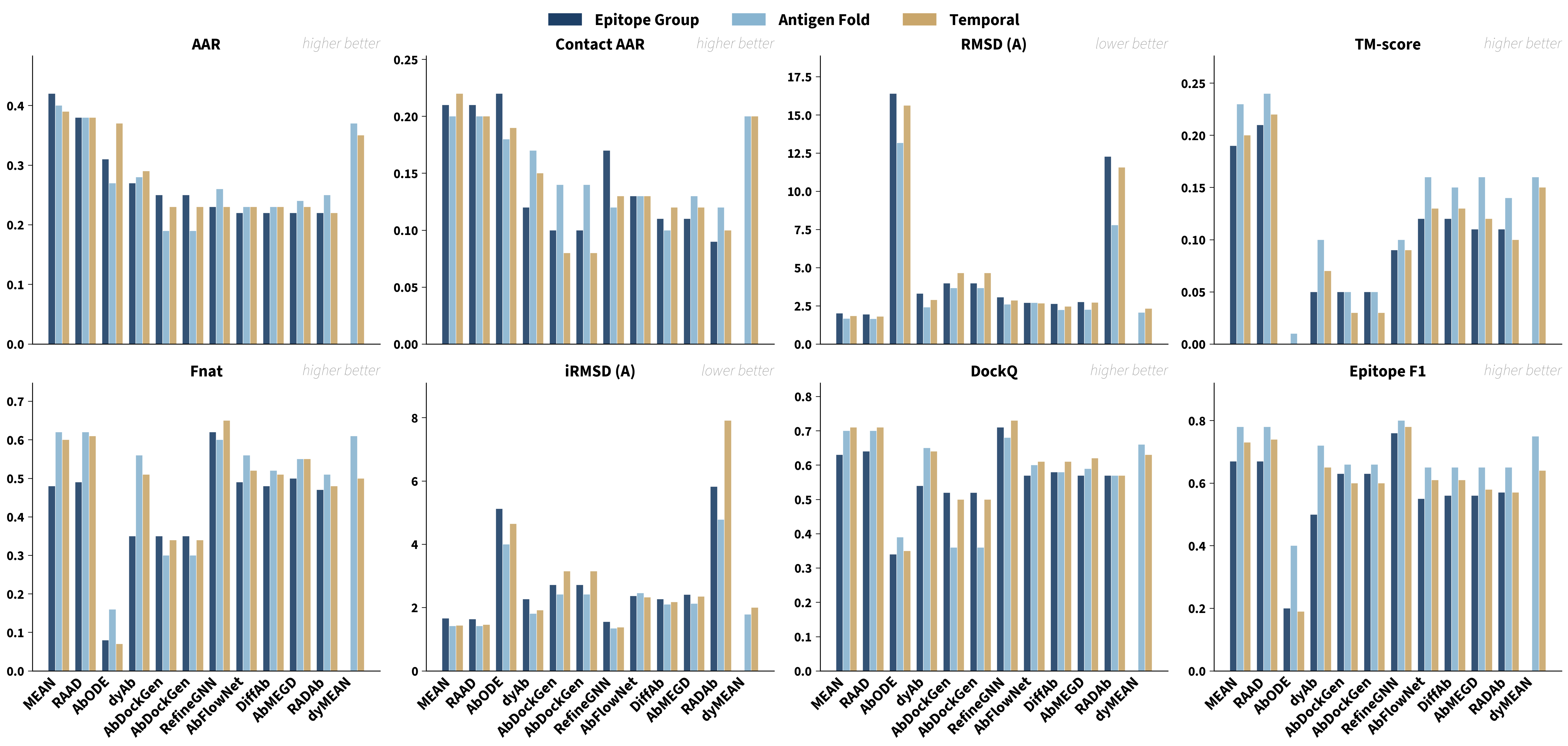}
\caption{CDR-H3 design metrics across all methods on the three test splits (EG: epitope-group, AF: antigen-fold, TP: temporal). }
\label{fig:leaderboards}
\end{figure}

Table~\ref{tab:results_epitope} reports CDR-H3 co-design results on the epitope-group test split, our primary evaluation setting since it holds out entire epitope clusters. Full results on the antigen-fold and temporal splits are deferred to Appendix~\ref{app:additional_results} (Tables~\ref{tab:results_antigen} and~\ref{tab:results_temporal}), and Figure~\ref{fig:leaderboards} visualizes the core metrics across all three splits. All values are computed using the standardized \chimera{} evaluation pipeline with Kabsch-aligned RMSD. Perplexity is reported as ``--'' for diffusion and flow-based models that do not produce sequence likelihoods. RADAb's large CDR-H3 RMSD and iRMSD with std larger than the mean reflect a minority of diffusion sampling failures that place the loop far off native.

\begin{table}[h!]
\centering
\caption{CDR-H3 design results on the \textbf{epitope-group} test split. Best values are in \textbf{bold}, second-best are \underline{underlined}. ``--'' indicates a metric that is not applicable (PPL for diffusion and flow models).  }
\label{tab:results_epitope}
\tiny
\setlength{\tabcolsep}{3pt}
\begin{tabular}{lllllllllll}
\toprule
Method & AAR$\uparrow$ & CAAR$\uparrow$ & PPL$\downarrow$ & RMSD$\downarrow$ & TM$\uparrow$ & Fnat$\uparrow$ & iRMSD$\downarrow$ & DockQ$\uparrow$ & EpiF1$\uparrow$ & n\_liab.$\downarrow$ \\
\midrule
RAAD & \underline{0.38$\pm$0.13} & \underline{0.21$\pm$0.23} & \underline{3.31$\pm$0.47} & \textbf{1.95$\pm$0.87} & \textbf{0.21$\pm$0.13} & 0.49$\pm$0.30 & \underline{1.64$\pm$0.86} & \underline{0.64$\pm$0.21} & \underline{0.67$\pm$0.27} & 0.71$\pm$0.62 \\
MEAN & \textbf{0.42$\pm$0.15} & \underline{0.21$\pm$0.23} & \textbf{3.00$\pm$0.61} & \underline{2.01$\pm$0.89} & \underline{0.19$\pm$0.12} & 0.48$\pm$0.31 & 1.66$\pm$0.86 & 0.63$\pm$0.21 & \underline{0.67$\pm$0.29} & 0.90$\pm$0.92 \\
dyMEAN & 0.35$\pm$0.12 & 0.17$\pm$0.20 & 3.38$\pm$0.46 & 2.60$\pm$0.94 & 0.12$\pm$0.07 & 0.41$\pm$0.30 & 2.23$\pm$0.84 & 0.56$\pm$0.19 & 0.55$\pm$0.29 & 0.83$\pm$0.38 \\
DiffAb & 0.22$\pm$0.12 & 0.11$\pm$0.18 & -- & 2.64$\pm$1.19 & 0.12$\pm$0.12 & 0.48$\pm$0.32 & 2.27$\pm$0.98 & 0.58$\pm$0.23 & 0.56$\pm$0.27 & 0.62$\pm$0.77 \\
AbFlowNet & 0.22$\pm$0.12 & 0.13$\pm$0.18 & -- & 2.70$\pm$1.24 & 0.12$\pm$0.12 & 0.49$\pm$0.33 & 2.37$\pm$1.01 & 0.57$\pm$0.22 & 0.55$\pm$0.28 & \underline{0.46$\pm$0.68} \\
AbMEGD & 0.22$\pm$0.11 & 0.11$\pm$0.18 & -- & 2.76$\pm$1.26 & 0.11$\pm$0.11 & \underline{0.50$\pm$0.32} & 2.41$\pm$1.04 & 0.57$\pm$0.22 & 0.56$\pm$0.27 & 0.47$\pm$0.72 \\
RADAb & 0.22$\pm$0.12 & 0.09$\pm$0.16 & -- & 12.28$\pm$75.62 & 0.11$\pm$0.10 & 0.47$\pm$0.32 & 5.82$\pm$31.86 & 0.57$\pm$0.22 & 0.57$\pm$0.28 & 0.83$\pm$0.90 \\
dyAb & 0.27$\pm$0.10 & 0.12$\pm$0.17 & -- & 3.31$\pm$0.84 & 0.05$\pm$0.03 & 0.35$\pm$0.28 & 2.27$\pm$0.86 & 0.54$\pm$0.18 & 0.50$\pm$0.32 & 0.98$\pm$0.15 \\
RefineGNN & 0.23$\pm$0.12 & 0.17$\pm$0.23 & 7.80$\pm$2.73 & 3.07$\pm$0.72 & 0.09$\pm$0.06 & \textbf{0.62$\pm$0.20} & \textbf{1.55$\pm$0.58} & \textbf{0.71$\pm$0.09} & \textbf{0.76$\pm$0.10} & 0.81$\pm$0.68 \\
AbDockGen & 0.23$\pm$0.11 & 0.09$\pm$0.16 & 7.89$\pm$2.94 & 3.93$\pm$1.26 & 0.05$\pm$0.03 & 0.34$\pm$0.25 & 2.64$\pm$1.03 & 0.52$\pm$0.17 & 0.62$\pm$0.27 & 0.60$\pm$0.74 \\
AbODE & 0.31$\pm$0.13 & \textbf{0.22$\pm$0.21} & 9.99$\pm$3.50 & 16.40$\pm$4.67 & 0.00$\pm$0.00 & 0.08$\pm$0.12 & 5.12$\pm$2.24 & 0.34$\pm$0.08 & 0.20$\pm$0.19 & \textbf{0.08$\pm$0.27} \\
\bottomrule
\end{tabular}
\end{table}

\subsection{All-CDR Design Results}
\label{sec:allcdr}

Beyond CDR-H3, several methods design all six CDR loops simultaneously. Table~\ref{tab:results_allcdr} reports per-CDR AAR, RMSD, and EpiF1 on the epitope-group split. Only DiffAb, AbFlowNet, AbMEGD, RADAb, and dyAb produce all six loops on this split, while the remaining methods are limited to the heavy-chain loops. A clear difficulty hierarchy emerges across CDR types, where H3 is hardest (lowest AAR, highest RMSD) and H1 is easiest.

\begin{table}[h!]
\centering
\caption{Per-CDR results on the \textbf{epitope-group} test split. Mean AAR, RMSD, and EpiF1 are shown per CDR type. ``--'' indicates the method does not support the given CDR. Best values are in \textbf{bold}, second-best are \underline{underlined}.}
\label{tab:results_allcdr}
\small
\setlength{\tabcolsep}{2pt}
\begin{tabular}{l|cccccc|cccccc|cccccc}
\toprule
 & \multicolumn{6}{c|}{AAR$\uparrow$} & \multicolumn{6}{c|}{RMSD$\downarrow$} & \multicolumn{6}{c}{EpiF1$\uparrow$} \\
Method & H1 & H2 & H3 & L1 & L2 & L3 & H1 & H2 & H3 & L1 & L2 & L3 & H1 & H2 & H3 & L1 & L2 & L3 \\
\midrule
RAAD      & \textbf{0.72} & \textbf{0.67} & \underline{0.38} & -- & -- & -- & \textbf{0.59} & \textbf{0.51} & \textbf{1.95} & -- & -- & -- & \textbf{0.92} & \textbf{0.89} & \underline{0.67} & -- & -- & -- \\
MEAN      & \underline{0.70} & 0.59 & \textbf{0.42} & -- & -- & -- & 0.85 & 0.83 & \underline{2.01} & -- & -- & -- & 0.86 & \underline{0.85} & 0.67 & -- & -- & -- \\
dyMEAN    & 0.67 & \underline{0.65} & 0.35 & -- & -- & -- & 1.07 & 0.99 & 2.60 & -- & -- & -- & 0.80 & 0.79 & 0.55 & -- & -- & -- \\
DiffAb    & 0.57 & 0.29 & 0.22 & \underline{0.55} & \underline{0.50} & \textbf{0.43} & 0.87 & 0.67 & 2.64 & 1.03 & 2.03 & \textbf{0.96} & 0.83 & 0.84 & 0.56 & 0.80 & \textbf{0.93} & 0.66 \\
AbFlowNet & 0.56 & 0.31 & 0.22 & 0.54 & \textbf{0.51} & 0.40 & 0.79 & \underline{0.65} & 2.70 & \textbf{0.99} & \underline{0.49} & \underline{0.96} & 0.84 & 0.84 & 0.55 & \textbf{0.82} & \underline{0.92} & 0.65 \\
AbMEGD    & 0.56 & 0.33 & 0.22 & 0.52 & 0.47 & 0.38 & 0.85 & 0.71 & 2.76 & \underline{1.00} & \textbf{0.48} & 1.09 & 0.82 & 0.83 & 0.56 & \underline{0.82} & 0.92 & 0.64 \\
RADAb     & 0.61 & 0.34 & 0.22 & \textbf{0.64} & \underline{0.50} & \underline{0.42} & 0.95 & 0.81 & 12.28 & 4.50 & 1.49 & 1.10 & 0.83 & 0.83 & 0.57 & 0.79 & 0.91 & \underline{0.67} \\
dyAb      & 0.48 & 0.32 & 0.27 & 0.24 & 0.46 & 0.26 & 1.87 & 1.82 & 3.31 & 1.64 & 1.11 & 1.37 & 0.75 & 0.72 & 0.50 & 0.81 & 0.89 & \textbf{0.83} \\
RefineGNN & 0.67 & 0.47 & 0.23 & -- & -- & -- & \underline{0.63} & 1.74 & 3.07 & -- & -- & -- & \underline{0.92} & 0.84 & \textbf{0.76} & -- & -- & -- \\
AbDockGen & -- & -- & 0.23 & -- & -- & -- & -- & -- & 3.93 & -- & -- & -- & -- & -- & 0.62 & -- & -- & -- \\
AbODE     & 0.53 & 0.56 & 0.31 & -- & -- & -- & 4.76 & 6.69 & 16.40 & -- & -- & -- & 0.45 & 0.54 & 0.20 & -- & -- & -- \\
\bottomrule
\end{tabular}
\end{table}

\subsection{Analysis}
\label{sec:analysis}

\paragraph{Sequence recovery and structural quality decouple.}
Equivariant GNN methods (RAAD, MEAN, dyMEAN) lead sequence recovery on the epitope-group split, with CDR-H3 AAR between 0.35 and 0.42, well above the diffusion and flow-matching methods (DiffAb, AbMEGD, AbFlowNet, dyAb) that cluster between 0.22 and 0.27. Structural accuracy follows a different ordering. RAAD and MEAN also produce the most accurate CDR-H3 backbones (RMSD near 2.0~\AA), whereas AbDockGen is the weakest method that still returns well-formed loops (RMSD=3.93~\AA). AbODE is a degenerate case, with CDR-H3 RMSD of 16.40~\AA{} and TM-score near zero across all three splits, a failure we trace to its polar-coordinate loop parameterization (\S\ref{sec:qualitative}). Notably, AbODE still reaches the highest contact-residue recovery on this split (CAAR=0.22), so its sequence predictions are reasonable even though its coordinates are not, a clear decoupling of sequence and structure.

\paragraph{Interface quality does not track sequence recovery.}
The interface metrics reorder the methods again. RefineGNN attains the best native-contact recovery and docking quality (Fnat=0.62, DockQ=0.71, EpiF1=0.76 on the epitope-group split) despite modest sequence recovery (AAR=0.23), while RAAD and MEAN trail on Fnat (0.48--0.49) even though they lead on AAR. At the other end, dyAb and AbODE recover few native contacts (Fnat=0.35 and 0.08). Sequence recovery alone is therefore a poor proxy for binding-interface quality, which motivates reporting the two metric groups separately.

\paragraph{Novel epitopes expose a generalization gap.}
A single random split would hide where these methods break. Sequence recovery is nearly invariant to the split, for example RAAD reaches AAR=0.38 on all three, but interface and epitope metrics degrade sharply on the epitope-group split, which holds out entire epitope clusters. RAAD's Fnat falls from 0.62 on antigen-fold to 0.49 on epitope-group, and its EpiF1 falls from 0.78 to 0.67. The same direction of change holds for nearly every method (Tables~\ref{tab:results_antigen} and~\ref{tab:results_temporal}), and Table~\ref{tab:gap} summarizes this epitope-group versus antigen-fold gap side by side. Binding-interface quality, not sequence recovery, is the axis that separates fitting seen epitopes from generalizing to novel ones, which is exactly what the epitope-group split and the epitope-specificity metrics are designed to expose.

\begin{figure}[h!]
\centering
\includegraphics[width=\textwidth]{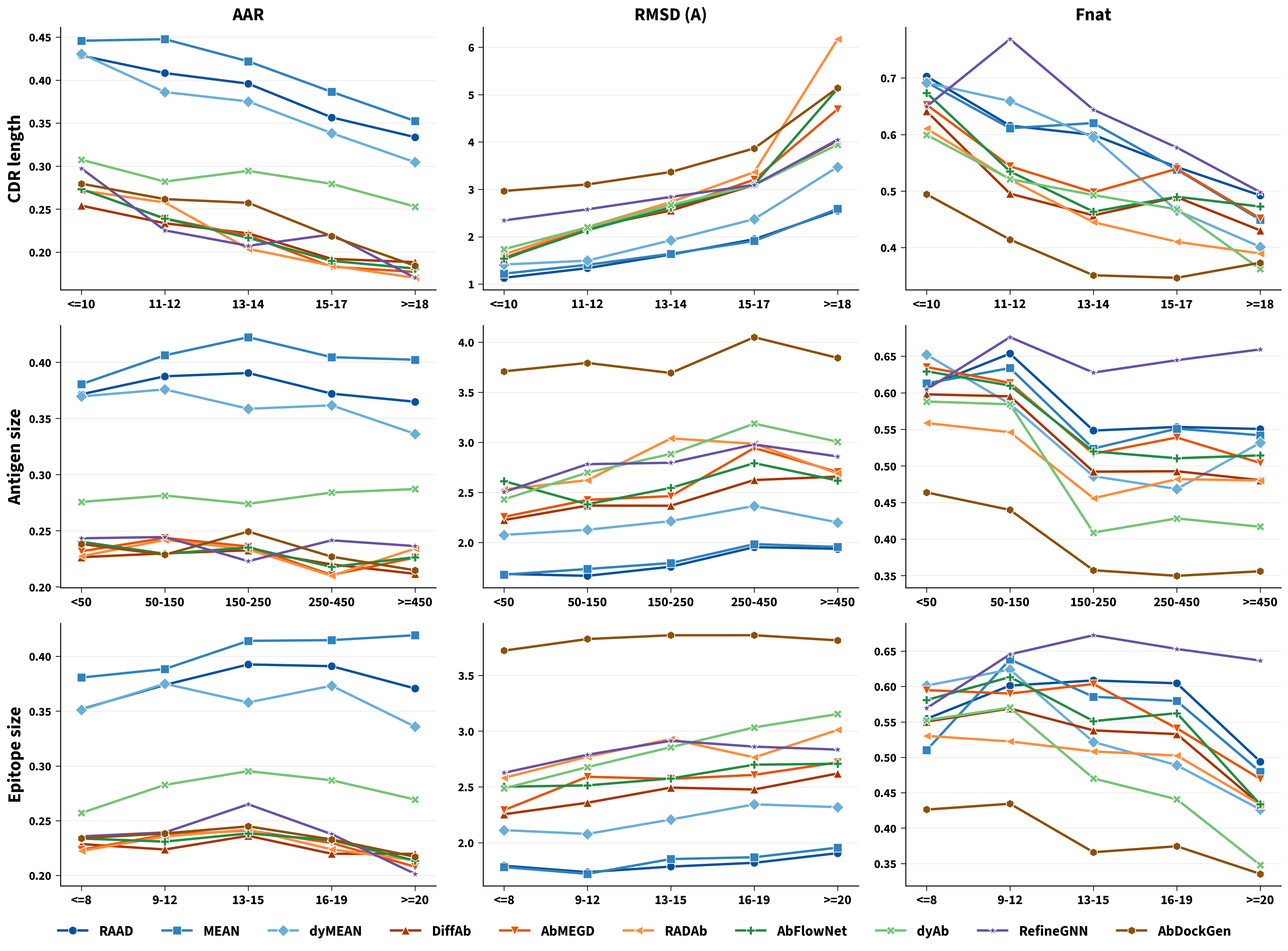}
\caption{CDR-H3 design quality on the epitope-group split, stratified by three covariates (rows: CDR-H3 length, antigen size, epitope size) across three metrics (columns: sequence recovery AAR, backbone accuracy RMSD, native-contact recovery Fnat). Loop length is the dominant difficulty axis, with all three metrics degrading as the loop lengthens, while antigen and epitope size leave AAR nearly flat.}
\label{fig:h3_strat_length}
\end{figure}

\paragraph{Difficulty scales with CDR-H3 length.}
Loop length is the dominant difficulty axis. The top row of Figure~\ref{fig:h3_strat_length} stratifies the epitope-group test set by CDR-H3 length and shows that sequence recovery, backbone accuracy, and native-contact recovery all degrade as the loop grows. The methods separate by how much. The one-shot denoising and equivariant-translation methods lose the most native-contact recovery from short to long loops, while RefineGNN degrades least and leads the long-loop regime despite its modest sequence recovery. Its autoregressive, structure-conditioned formulation appears to preserve binding contacts as the loop extends. By contrast, neither antigen size nor epitope size moves AAR appreciably (middle and bottom rows of Figure~\ref{fig:h3_strat_length}), so long CDR-H3 loops are the regime where new methods stand to gain the most.

\paragraph{Sequence recovery is insensitive to antigen and epitope properties.}
We further stratify the epitope-group CDR-H3 results by antigen size, epitope size, and antigen type. Sequence recovery (AAR) is insensitive to all three. CDR-H3 AAR is nearly flat across antigen sizes and epitope sizes (middle and bottom rows of Figure~\ref{fig:h3_strat_length}) and varies by at most 0.03 between protein and peptide antigens. The interface metrics are the ones that move. Native-contact recovery is non-monotonic in epitope size, peaking on medium epitopes of 7 to 15 residues and dropping on both smaller epitopes, which offer too few contacts to recover, and larger epitopes, which offer too many to place correctly (bottom row of Figure~\ref{fig:h3_strat_length}). Peptide antigens are systematically easier on the interface, with most methods gaining 0.03 to 0.15 Fnat over protein antigens, so the protein-antigen subset is the harder and more informative slice. As in the length analysis, RefineGNN is the most robust to these covariates on the interface metrics.

\paragraph{Capacity and compute do not predict quality.}
Larger models and longer training do not yield better designs on \chimera{}. The smallest trainable model, MEAN at 0.7M parameters, matches or beats every diffusion model on CDR-H3 across all three splits, including RADAb, which adds a frozen 650M-parameter ESM-2 encoder and a retrieval stage for no measurable gain over plain DiffAb. The gap is more prominent in cost-adjusted terms. MEAN and RAAD train in under 4 GPU-hours in total, and yet match or beat the diffusion models that each need 6 to 11 hours (Table~\ref{tab:compute_detail}). Inference cost spans three orders of magnitude, from under a minute per split for the lightest GNNs to roughly two to three hours per split for RADAb, whose MSA construction and retrieval stage dominate its runtime. Our analysis shows that architecture and inductive bias, not capacity, drive performance in this benchmark.

\begin{table}[h!]
\centering
\caption{Computational resources for all evaluated methods. Params are total and trainable parameter counts. Train is total wall-clock training time in GPU-hours summed over all CDR models a method trains, and Infer is total test-set inference time in minutes, both reported per split (epitope group: EG, antigen fold: AF, temporal: TP).  }
\label{tab:compute_detail}
\small
\setlength{\tabcolsep}{4pt}
\begin{tabular}{llrrcccccc}
\toprule
 & & \multicolumn{2}{c}{Params} & \multicolumn{3}{c}{Train (GPU-h)} & \multicolumn{3}{c}{Infer (min)} \\
\cmidrule(lr){3-4}\cmidrule(lr){5-7}\cmidrule(lr){8-10}
Method & Type & Total & Train. & EG & AF & TP & EG & AF & TP \\
\midrule
MEAN & Equiv.\ GNN & 0.7M & 0.7M & 3.5 & 2.6 & 2.4 & 0.5 & 0.5 & 0.6 \\
RAAD & Equiv.\ GNN & 6.8M & 6.8M &  2.8 & 3.0 & 2.1 & 0.8 & 0.7 & 0.5 \\
dyMEAN & Equiv.\ GNN & 2.0M & 2.0M  & 15.3 & 14.5 & 16.5 & 0.7 & 1.4 & 0.6 \\
RefineGNN & Autoreg.\ GNN & 5.8M & 5.8M & 3.5 & 3.2 & 3.1 & 2.8 & 3.2 & 3.2 \\
dyAb & Flow matching & 2.5M & 2.5M &  1.1 & 0.7 & 0.7 & 2.2 & 1.6 & 1.6 \\
AbDockGen & Hier.\ ENN & 8.8M & 8.8M  & 2.3 & 1.6 & 1.6 & 2.9 & 1.7 & 2.0 \\
AbODE & Neural ODE & 0.24M & 0.24M &  13.9 & 7.0 & 9.0 & 0.9 & 0.9 & 0.9 \\
DiffAb & Diffusion & 4.0M & 4.0M &  7.4 & 6.0 & 6.6 & 8.5 & 7.9 & 10.0 \\
AbFlowNet & Flow matching & 4.0M & 4.0M  & 8.3 & 4.7 & 6.1 & 6.6 & 10.0 & 11.0 \\
AbMEGD & Diffusion & 5.8M & 5.8M & 9.8 & 10.2 & 11.2 & 7.8 & 7.6 & 7.8 \\
RADAb & Retr.+Diff. & 661M & 10.1M  & 6.5 & 7.4 & 10.7 & 119.6 & 113.2 & 204.4 \\
\bottomrule
\end{tabular}
\end{table}

\paragraph{No single method dominates.}
The best method depends on both the split and the metric group. RAAD and MEAN lead sequence recovery and backbone accuracy on every split, while RefineGNN leads the interface and epitope metrics on the held-out epitope-group and temporal splits. No method wins more than four of the eight CDR-H3 metrics on any one split, which is why we report each split and metric group separately rather than collapsing them into a single ranking. The practical consequence is clear. The open problem is generalization to novel epitopes, not raw sequence recovery, and the highest-value regime for new methods is long CDR-H3 loops evaluated with interface metrics, where structure-conditioned autoregressive inductive biases outperform the largest models.

\subsection{Qualitative CDR-H3 Structure Comparison}
\label{sec:qualitative}

\begin{figure}[h!]
    \centering
    \includegraphics[width=0.75\linewidth]{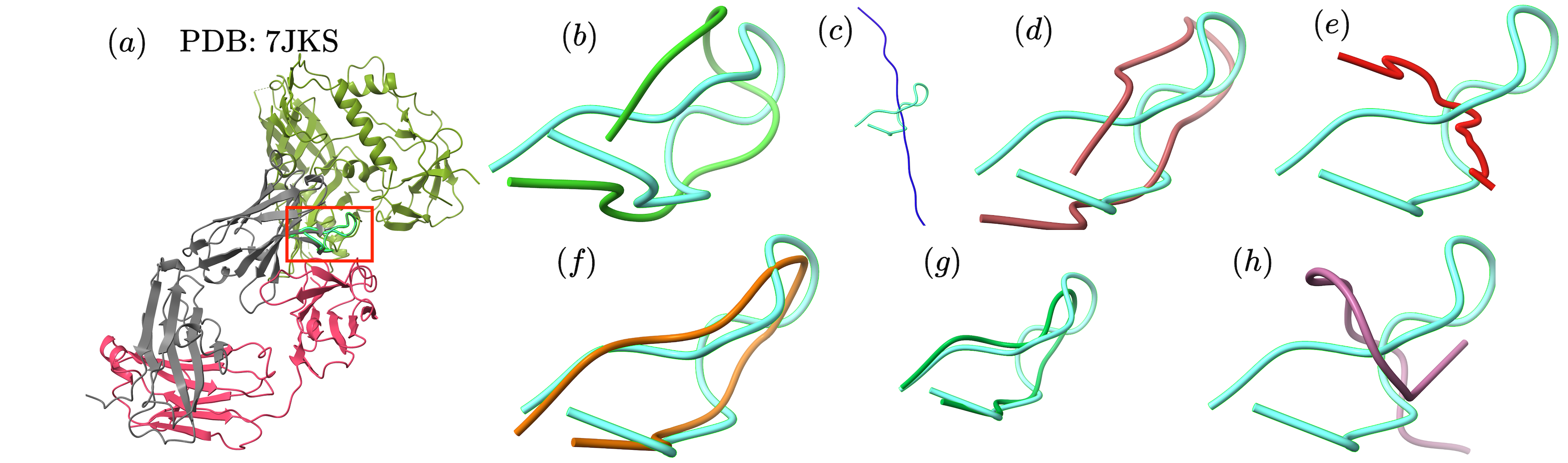} \vskip.2in
    \includegraphics[width=0.75\linewidth]{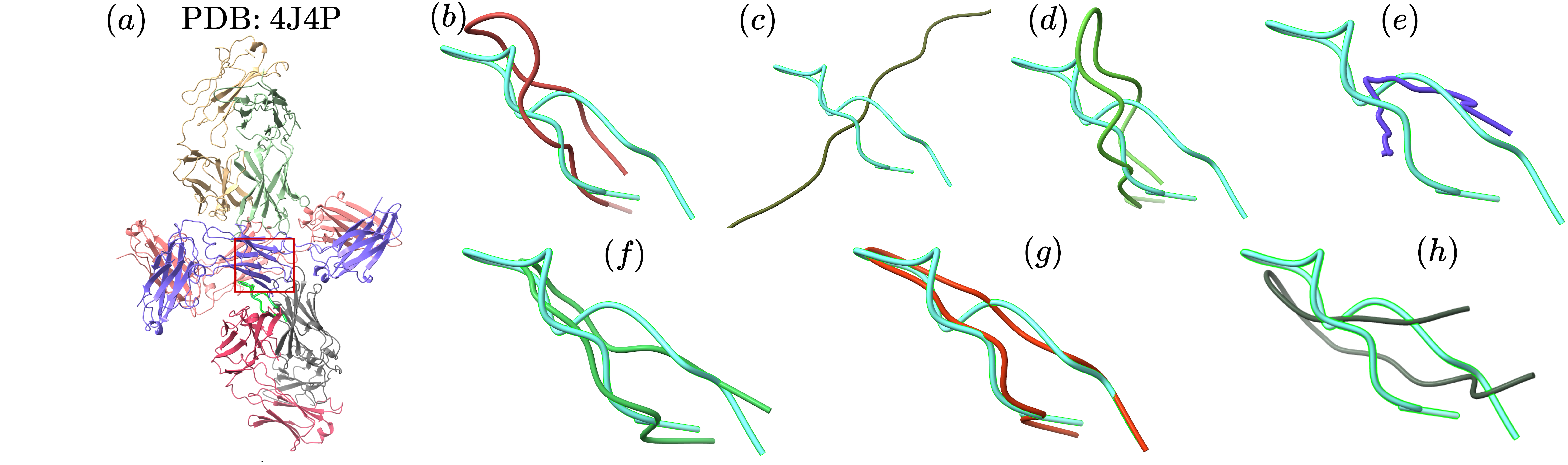}  \vskip.2in
    \includegraphics[width=0.75\linewidth]{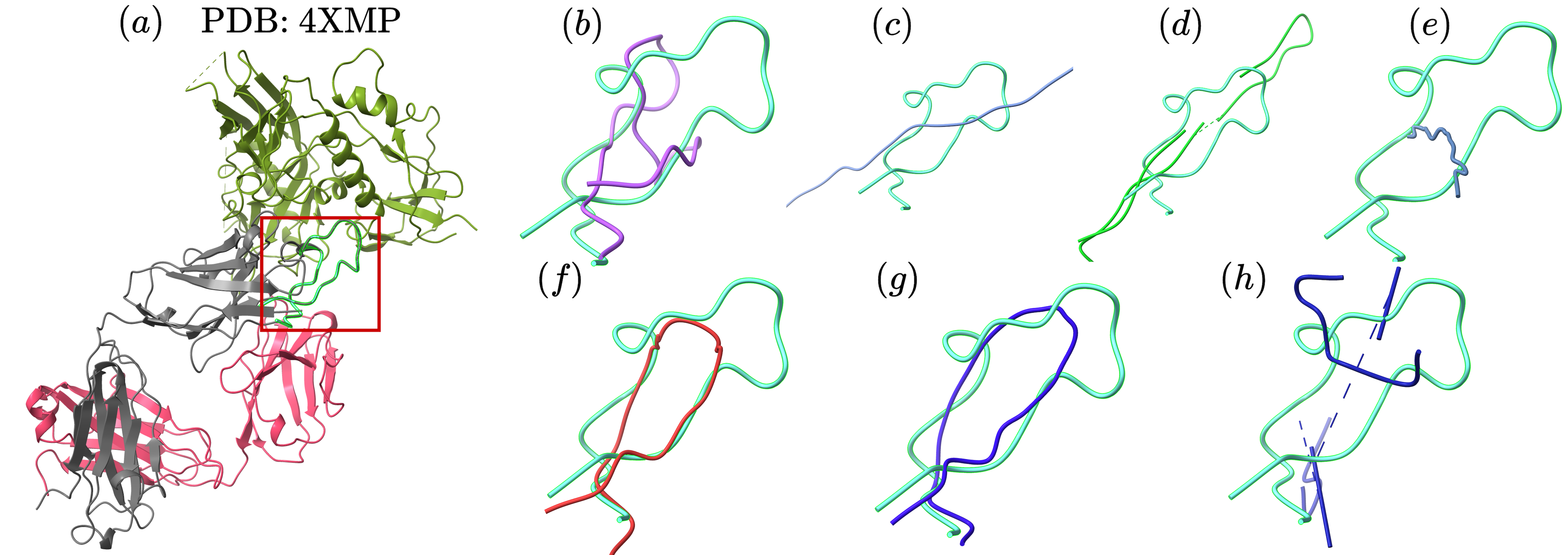}
    \caption{Native (a) and predicted CDR-H3 backbone structures for three complexes of increasing difficulty: 7JKS (H3 length 14), 4J4P (H3 length 17), and 4XMP (H3 length 25). Panels show predictions from (b) AbFlowNet, (c) AbODE, (d) DiffAb, (e) dyAb, (f) dyMEAN, (g) RAAD, (h) RefineGNN. The native CDR H3 regions are shown in cyan colors.}
    \label{fig:h3_structures}
\end{figure}

Figure~\ref{fig:h3_structures} shows the predicted CDR-H3 backbone structures from seven methods overlaid on the native loop for three complexes selected to illustrate different design regimes.
\textbf{7JKS} (H3 length 14) is a typical case where equivariant GNN methods (RAAD, dyMEAN) recover the native sequence well (AAR${\sim}$0.36) with moderate RMSD (${\sim}$1.6~\AA), while diffusion-based methods achieve lower AAR (${\sim}$0.17--0.33) with slightly higher RMSD (${\sim}$2.1--2.7~\AA).
\textbf{4J4P} (H3 length 17) represents a case where equivariant GNNs and diffusion methods disagree. dyMEAN achieves the highest AAR (0.71) while diffusion methods recover substantially fewer residues (AAR${\sim}$0.13--0.27), illustrating how longer loops amplify both sequence and structural prediction errors.
\textbf{4XMP} (H3 length 25) is a challenging long-loop case where all methods struggle. AAR drops below 0.18 for every method, and the diffusion-based methods produce particularly large structural deviations on this long loop. AbODE returns a degenerate global placement here, consistent with its catastrophic aggregate RMSD in the main results.

\subsection{Limitations}
\label{sec:limitations}

\chimera{} evaluates designs against static crystallographic structures and does not model conformational flexibility, solvent effects, or experimental binding affinity, so high benchmark scores are necessary but not sufficient evidence of a viable binder, and the dataset inherits the biases of SAbDab and the PDB toward viral surface antigens, human-derived antibodies, and X-ray crystallography (66\% of complexes). The coverage is also incomplete, as several methods could not be evaluated under leakage control, including RFAntibody~\citep{bennett2025rfdiffusion-ab}, whose weights are trained on the full PDB without a disclosed split, and ProteinMPNN~\citep{dauparas2022proteinmpnn}, whose inverse-folding benchmarking could be the future work.

\section{Conclusion}
\label{sec:conclusion}

We have presented \chimera{}, a unified benchmark for epitope-specific antibody CDR design. \chimera{} provides a canonical task formulation, a curated dataset of 2,922 complexes with biologically motivated splits, and a standardized evaluation protocol, including novel epitope-specificity measures. Our evaluation of eleven baseline methods spanning different generative paradigms establishes comparisons under consistent conditions and reveals patterns invisible under a single random split. 

\subsubsection*{Broader Impact Statement}
In this optional section, TMLR encourages authors to discuss possible repercussions of their work,
notably any potential negative impact that a user of this research should be aware of. 
Authors should consult the TMLR Ethics Guidelines available on the TMLR website
for guidance on how to approach this subject.


\subsubsection*{Acknowledgments}

The work used the Jetstream2 Supercomputer at Indiana University through allocation CIS251200 from the Advanced Cyberinfrastructure Coordination Ecosystem: Services \& Support (ACCESS) program~\citep{boerner2023access}, which is supported by National Science Foundation grants \#2138259, \#2138286, \#2138307, \#2137603, and \#2138296. 

\bibliography{references}
\bibliographystyle{tmlr}

\newpage

\appendix
\section{Appendix}

\subsection{Processing Details}
\label{app:processing}

The six sequential quality filters applied during dataset construction are: (1) restriction to protein and peptide antigens, excluding carbohydrates, nucleic acids, and haptens, (2) requirement for both heavy (VH) and light (VL) chains, (3) requirement for an annotated antigen chain, (4) a crystallographic resolution cutoff of 4.0~\AA, (5) verification that the structure file was successfully downloaded, and (6) removal of duplicate rows sharing the same PDB, heavy chain, light chain, and antigen chain identifiers. Sequence deduplication concatenates VH and VL sequences for each complex and clusters at 95\% identity with 80\% coverage using MMseqs2~\citep{steinegger2017mmseqs2}, retaining the highest-resolution representative per cluster, yielding 2,981 deduplicated complexes.

The validation step verifies baseline compatibility by checking four criteria: successful ANARCI numbering for both VH and VL chains, presence of conserved residues at IMGT positions 23 (Cys), 41 (Trp), and 104 (Cys), identifiability of all six CDR regions under IMGT numbering, and no missing C$\alpha$ atoms in any chain. This excludes 59 complexes, of which 27 have missing backbone atoms (mostly peptide antigen termini), 12 contain atypical V-genes that ANARCI cannot number, 11 have engineered conserved-residue substitutions, and 9 have incomplete CDR or conserved-position annotations (including 3 lambda chains lacking CDR-L2 and 6 with missing IMGT anchor positions or CDR regions). The excluded complexes and their reasons are recorded for transparency.


\subsection{Metric Definitions}
\label{app:metrics}

We provide formal definitions for all metrics used in \chimera{}. Let $\hat{s} = (\hat{s}_1, \ldots, \hat{s}_N)$ and $s = (s_1, \ldots, s_N)$ denote the predicted and native CDR sequences of length $N$, $\hat{\mathbf{X}}, \mathbf{X} \in \mathbb{R}^{N \times 3}$ the predicted and native C$\alpha$ coordinates, $m \in \{0,1\}^N$ the paratope contact mask, and $\mathcal{C}, \hat{\mathcal{C}}$ the native and predicted sets of antibody--antigen contact pairs. We group these metrics into five categories as follows:

\subsubsection{Sequence quality.}
These metrics measure how well a design recovers the native CDR sequence.

\textbf{Amino Acid Recovery (AAR)} is the fraction of CDR positions whose predicted residue matches the native residue. It ranges from 0 to 1, and higher is better.
\begin{equation}
    \text{AAR} = \frac{1}{N} \sum_{i=1}^{N} \mathbb{1}[\hat{s}_i = s_i]
\end{equation}

\textbf{Contact Amino Acid Recovery (CAAR)} restricts AAR to the paratope contact positions, isolating recovery at the binding-relevant residues that drive affinity. It ranges from 0 to 1, and higher is better.
\begin{equation}
    \text{CAAR} = \frac{\sum_{i=1}^{N} m_i \cdot \mathbb{1}[\hat{s}_i = s_i]}{\sum_{i=1}^{N} m_i}
\end{equation}

\textbf{Perplexity (PPL)} is the exponentiated per-residue negative log-likelihood the model assigns to the native sequence, measuring how confidently and fluently a model reproduces it. Lower is better. It is defined only for likelihood-based models and is reported as not applicable for diffusion and flow models.
\begin{equation}
    \text{PPL} = \exp\!\left(-\frac{1}{N}\sum_{i=1}^{N} \log p_\theta(s_i \mid \text{context})\right)
\end{equation}

\subsubsection{Structural quality.}
These metrics measure how well a design reproduces the native CDR backbone geometry.

\textbf{Root Mean Square Deviation (RMSD)} is the average C$\alpha$ distance between the predicted and native loop after optimal SVD-based Kabsch superposition, reported in \AA{} with lower being better. Here $R^*$ is the optimal rotation from the SVD of $\hat{\mathbf{X}}_c^\top \mathbf{X}_c$ on centered coordinates.
\begin{equation}
    \text{RMSD} = \sqrt{\frac{1}{N} \sum_{i=1}^{N} \| R^* \hat{\mathbf{x}}_{c,i} - \mathbf{x}_{c,i} \|^2}
\end{equation}

\textbf{Template Modeling score (TM)}~\citep{zhang2004tm} is a length-normalized measure of global fold similarity on the Kabsch-aligned structures, less sensitive to local outliers than RMSD. It ranges from 0 to 1, and higher is better. Here $d_i$ is the distance between aligned C$\alpha$ atoms and $L$ is the target length.
\begin{equation}
    \text{TM} = \frac{1}{L} \sum_{i=1}^{L} \frac{1}{1 + (d_i / d_0)^2}, \quad d_0 = \max\!\bigl(1.24\sqrt[3]{L-15} - 1.8,\; 0.5\bigr)
\end{equation}

\subsubsection{Binding interface quality.}
These metrics measure how well a design reproduces the native antibody--antigen interface. Two residues in different chains interact if their C$_\alpha$ atoms are within 8~\AA~\citep{xue2015computational,ovchinnikov2014fnat}. We follow this definition as a proxy for docking evaluation by CAPRI (Critical Assessment of Predicted Interactions)~\cite{janin2003capri} since some baseline methods output side chains~\citep{luo2022diffab,abir2025abflownet,wu2025raad,wang2024radab} while some other methods output only the backbone atoms~\citep{kong2023dymean,tan2025dyab,jin2021refinegnn,kong2022mean}.

\textbf{Fraction of Native Contacts (Fnat)} is the proportion of native antibody--antigen contact pairs that the design recovers. It ranges from 0 to 1, and higher is better.
\begin{equation}
    \text{Fnat} = \frac{|\hat{\mathcal{C}} \cap \mathcal{C}|}{|\mathcal{C}|}
\end{equation}

\textbf{Interface RMSD (iRMSD)} is the C$\alpha$ RMSD computed over the interface residues alone, in \AA{} with lower being better. $\mathbf{X}_\text{iface}$ denotes the C$\alpha$ coordinates of residues participating in contacts.
\begin{equation}
    \text{iRMSD} = \text{RMSD}\bigl(\hat{\mathbf{X}}_\text{iface},\, \mathbf{X}_\text{iface}\bigr)
\end{equation}

\textbf{Docking Quality (DockQ)}~\citep{basu2016dockq} is a single CAPRI-style score that combines Fnat, iRMSD, and the Kabsch-aligned full-antibody C$\alpha$ RMSD (LRMSD). It ranges from 0 to 1, and higher is better. Both native and predicted contact sets are computed symmetrically from C$\alpha$--C$\alpha$ distances at the 8~\AA{} cutoff.
\begin{equation}
    \text{DockQ} = \frac{1}{3}\left(\text{Fnat} + \frac{1}{1+(\text{iRMSD}/1.5)^2} + \frac{1}{1+(\text{LRMSD}/8.5)^2}\right)
\end{equation}

\subsubsection{Epitope specificity.}
These metrics measure whether the design contacts the intended binding site. Let $\hat{E}$ be the set of antigen residues contacted by the designed antibody (within 8~\AA{} C$\alpha$--C$\alpha$ distance) and $E$ the true epitope residues.

\textbf{Epitope Precision (EpiPrec)} is the fraction of contacted antigen residues that belong to the true epitope, and \textbf{Epitope Recall (EpiRec)} is the fraction of true epitope residues that the design contacts. \textbf{Epitope F1 (EpiF1)} is their harmonic mean and is the headline epitope metric. All three range from 0 to 1, and higher is better.
\begin{equation}
    \text{EpiPrec} = \frac{|\hat{E} \cap E|}{|\hat{E}|}, \quad
    \text{EpiRec} = \frac{|\hat{E} \cap E|}{|E|}, \quad
    \text{EpiF1} = \frac{2 \cdot \text{EpiPrec} \cdot \text{EpiRec}}{\text{EpiPrec} + \text{EpiRec}}
\end{equation}

\subsubsection{Designability.}
\textbf{Sequence liability count (n\_liab.)} is the number of known developability liability motifs (NG, DG, DS, DD, NS, NT, M) present in the designed CDR sequence. It is a non-negative integer, and lower is better.

\subsection{Pipeline Configuration}
\label{app:config}

Table~\ref{tab:config} lists the values of all thresholds and processing parameters used throughout this work. Table~\ref{tab:cdr_defs} provides the CDR boundary definitions under both numbering schemes.

\begin{table}[h!]
\centering
\caption{Pipeline configuration parameters.}
\label{tab:config}
\small
\begin{tabular}{llc}
\toprule
Parameter & Description & Value \\
\midrule
Max resolution & Structure quality threshold & 4.0~\AA \\
Contact cutoff & Epitope/paratope definition & 4.5~\AA \\
Sequence identity & MMseqs2 clustering threshold & 95\% \\
Coverage & MMseqs2 alignment coverage & 80\% \\
Numbering schemes & Residue numbering & IMGT, Chothia \\
Graph $k$-NN & Nearest neighbors for intra-chain edges & 10 \\
Graph spatial cutoff & Intra-chain spatial edge threshold & 8.0~\AA \\
Inter-chain cutoff & Inter-chain spatial edge threshold & 12.0~\AA \\
Train/Val/Test ratio & Split proportions & 80/10/10 \\
Temporal split & Date-based quantile cutoffs & 80/10/10 \\
\bottomrule
\end{tabular}
\end{table}

\begin{table}[h!]
\centering
\caption{CDR boundary definitions used in \chimera{}.}
\label{tab:cdr_defs}
\small
\begin{tabular}{lcccccc}
\toprule
Scheme & H1 & H2 & H3 & L1 & L2 & L3 \\
\midrule
IMGT~\citep{lefranc2003imgt}    & 27--38 & 56--65 & 105--117 & 27--38 & 56--65 & 105--117 \\
Chothia~\citep{chothia1987canonical} & 26--32 & 52--56 & 95--102 & 24--34 & 50--56 & 89--97 \\
\bottomrule
\end{tabular}
\end{table}

\subsection{Additional Results}
\label{app:additional_results}

\begin{figure}[h!]
\centering
\includegraphics[width=\textwidth]{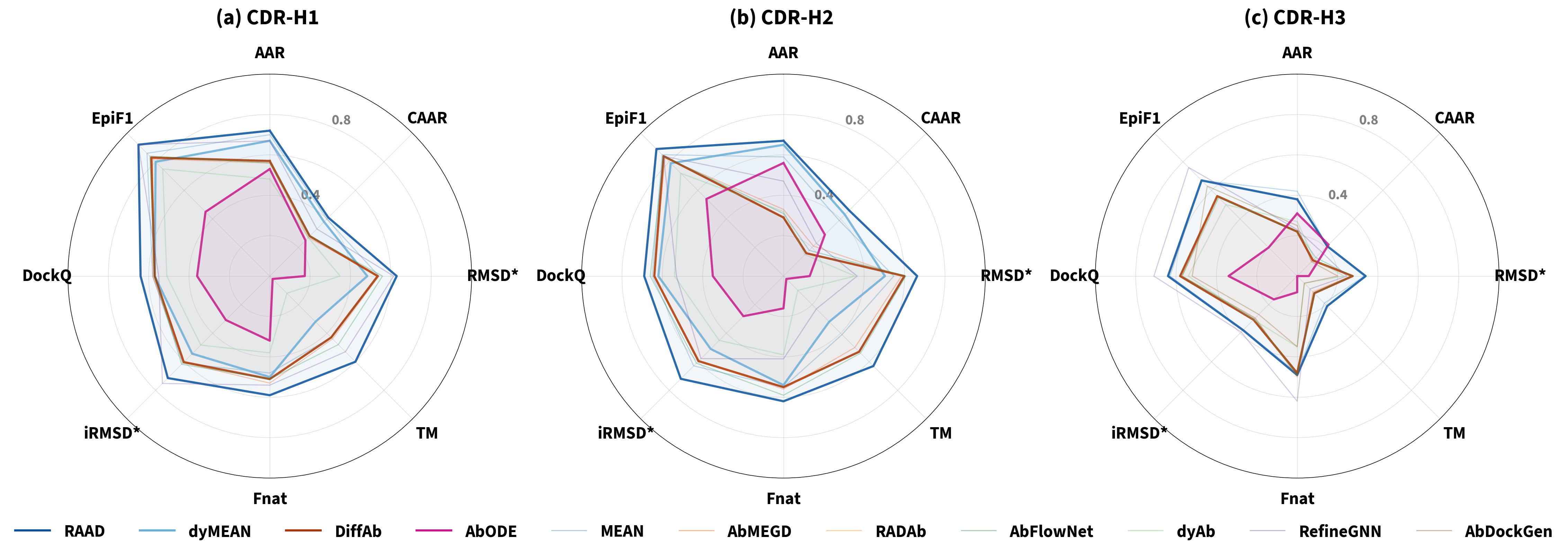}
\caption{Per-CDR comparison of methods across all evaluation metrics for (a) CDR-H1, (b) CDR-H2, and (c) CDR-H3. Each axis is scaled so that a larger radius is better; RMSD and iRMSD (marked with an asterisk) are inverted accordingly.}
\label{fig:radar}
\end{figure}

Tables~\ref{tab:results_antigen} and~\ref{tab:results_temporal} report the full CDR-H3 results on the antigen-fold and temporal test splits, complementing the epitope-group results in Table~\ref{tab:results_epitope}. Figure~\ref{fig:radar} compares the methods across all evaluation metrics for each heavy-chain CDR.

\begin{table}[h!]
\centering
\caption{CDR-H3 design results on the \textbf{antigen-fold} test split. Best values are in \textbf{bold}, second-best are \underline{underlined}. }
\label{tab:results_antigen}
\tiny
\setlength{\tabcolsep}{3pt}
\begin{tabular}{lllllllllll}
\toprule
Method & AAR$\uparrow$ & CAAR$\uparrow$ & PPL$\downarrow$ & RMSD$\downarrow$ & TM$\uparrow$ & Fnat$\uparrow$ & iRMSD$\downarrow$ & DockQ$\uparrow$ & EpiF1$\uparrow$ & n\_liab.$\downarrow$ \\
\midrule
RAAD~\citep{wu2025raad} & \underline{0.38$\pm$0.13} & \textbf{0.20$\pm$0.21} & \underline{3.06$\pm$0.49} & \textbf{1.65$\pm$0.70} & \textbf{0.24$\pm$0.13} & \textbf{0.62$\pm$0.31} & \underline{1.42$\pm$0.62} & \textbf{0.70$\pm$0.21} & \underline{0.78$\pm$0.25} & 0.59$\pm$0.60 \\
MEAN~\citep{kong2022mean} & \textbf{0.40$\pm$0.14} & \underline{0.20$\pm$0.23} & \textbf{2.79$\pm$0.57} & \underline{1.66$\pm$0.64} & \underline{0.23$\pm$0.11} & \underline{0.62$\pm$0.32} & 1.42$\pm$0.61 & \underline{0.70$\pm$0.21} & 0.78$\pm$0.26 & 0.38$\pm$0.70 \\
dyMEAN~\citep{kong2023dymean} & 0.37$\pm$0.13 & 0.20$\pm$0.23 & 3.35$\pm$0.66 & 2.06$\pm$0.83 & 0.16$\pm$0.09 & 0.61$\pm$0.31 & 1.79$\pm$0.82 & 0.66$\pm$0.20 & 0.75$\pm$0.25 & \underline{0.21$\pm$0.41} \\
DiffAb~\citep{luo2022diffab} & 0.23$\pm$0.12 & 0.10$\pm$0.15 & -- & 2.24$\pm$1.02 & 0.15$\pm$0.13 & 0.52$\pm$0.34 & 2.10$\pm$0.89 & 0.58$\pm$0.25 & 0.65$\pm$0.30 & 0.70$\pm$0.77 \\
AbFlowNet~\citep{abir2025abflownet} & 0.23$\pm$0.12 & 0.13$\pm$0.18 & -- & 2.70$\pm$5.36 & 0.16$\pm$0.14 & 0.56$\pm$0.34 & 2.46$\pm$4.31 & 0.60$\pm$0.26 & 0.65$\pm$0.30 & 0.61$\pm$0.81 \\
AbMEGD~\citep{chen2025AbMEGD} & 0.24$\pm$0.13 & 0.13$\pm$0.20 & -- & 2.25$\pm$1.06 & 0.16$\pm$0.15 & 0.55$\pm$0.34 & 2.13$\pm$0.94 & 0.59$\pm$0.25 & 0.65$\pm$0.30 & 0.64$\pm$0.76 \\
RADAb~\citep{wang2024radab} & 0.25$\pm$0.13 & 0.12$\pm$0.17 & -- & 7.79$\pm$41.92 & 0.14$\pm$0.13 & 0.51$\pm$0.33 & 4.78$\pm$22.78 & 0.57$\pm$0.26 & 0.65$\pm$0.31 & 0.57$\pm$0.79 \\
dyAb~\citep{tan2025dyab} & 0.28$\pm$0.11 & 0.17$\pm$0.21 & -- & 2.40$\pm$0.81 & 0.10$\pm$0.05 & 0.56$\pm$0.33 & 1.81$\pm$0.81 & 0.65$\pm$0.21 & 0.72$\pm$0.29 & 0.92$\pm$0.27 \\
RefineGNN~\citep{jin2021refinegnn} & 0.26$\pm$0.13 & 0.12$\pm$0.16 & 8.17$\pm$3.99 & 2.59$\pm$0.65 & 0.10$\pm$0.06 & 0.60$\pm$0.32 & \textbf{1.35$\pm$0.54} & 0.68$\pm$0.22 & \textbf{0.80$\pm$0.26} & 0.59$\pm$0.90 \\
AbDockGen~\citep{jin2022hern} & 0.24$\pm$0.13 & 0.10$\pm$0.16 & 7.89$\pm$3.45 & 3.65$\pm$0.81 & 0.05$\pm$0.03 & 0.46$\pm$0.27 & 2.41$\pm$0.84 & 0.57$\pm$0.17 & 0.75$\pm$0.23 & 0.59$\pm$0.68 \\
AbODE~\citep{verma2023abode} & 0.27$\pm$0.13 & 0.18$\pm$0.20 & 10.78$\pm$3.13 & 13.17$\pm$4.81 & 0.01$\pm$0.01 & 0.16$\pm$0.22 & 4.00$\pm$1.79 & 0.39$\pm$0.16 & 0.40$\pm$0.36 & \textbf{0.20$\pm$0.51} \\
\bottomrule
\end{tabular}
\end{table}

\begin{table}[h!]
\centering
\caption{CDR-H3 design results on the \textbf{temporal} test split. Best values are in \textbf{bold}, second-best are \underline{underlined}. }
\label{tab:results_temporal}
\tiny
\setlength{\tabcolsep}{3pt}
\begin{tabular}{lllllllllll}
\toprule
Method & AAR$\uparrow$ & CAAR$\uparrow$ & PPL$\downarrow$ & RMSD$\downarrow$ & TM$\uparrow$ & Fnat$\uparrow$ & iRMSD$\downarrow$ & DockQ$\uparrow$ & EpiF1$\uparrow$ & n\_liab.$\downarrow$ \\
\midrule
RAAD~\citep{wu2025raad} & \underline{0.38$\pm$0.13} & \underline{0.20$\pm$0.22} & \textbf{3.24$\pm$0.44} & \textbf{1.81$\pm$0.76} & \textbf{0.22$\pm$0.12} & \underline{0.61$\pm$0.28} & 1.46$\pm$0.70 & \underline{0.71$\pm$0.16} & \underline{0.74$\pm$0.23} & 0.77$\pm$0.65 \\
MEAN~\citep{kong2022mean} & \textbf{0.39$\pm$0.13} & \textbf{0.22$\pm$0.23} & 3.31$\pm$0.53 & \underline{1.84$\pm$0.68} & \underline{0.20$\pm$0.09} & 0.60$\pm$0.29 & \underline{1.44$\pm$0.66} & 0.71$\pm$0.16 & 0.73$\pm$0.23 & 0.66$\pm$0.66 \\
dyMEAN~\citep{kong2023dymean} & 0.35$\pm$0.13 & 0.20$\pm$0.22 & \underline{3.28$\pm$0.42} & 2.32$\pm$0.97 & 0.15$\pm$0.10 & 0.50$\pm$0.30 & 2.00$\pm$0.92 & 0.63$\pm$0.17 & 0.64$\pm$0.26 & 0.99$\pm$0.71 \\
DiffAb~\citep{luo2022diffab} & 0.23$\pm$0.11 & 0.12$\pm$0.17 & -- & 2.46$\pm$1.05 & 0.13$\pm$0.10 & 0.51$\pm$0.30 & 2.18$\pm$0.95 & 0.61$\pm$0.19 & 0.61$\pm$0.26 & 0.74$\pm$0.84 \\
AbFlowNet~\citep{abir2025abflownet} & 0.23$\pm$0.12 & 0.13$\pm$0.19 & -- & 2.67$\pm$2.79 & 0.13$\pm$0.11 & 0.52$\pm$0.32 & 2.33$\pm$1.98 & 0.61$\pm$0.20 & 0.61$\pm$0.26 & 0.45$\pm$0.70 \\
AbMEGD~\citep{chen2025AbMEGD} & 0.23$\pm$0.11 & 0.12$\pm$0.17 & -- & 2.72$\pm$2.42 & 0.12$\pm$0.10 & 0.55$\pm$0.31 & 2.35$\pm$1.67 & 0.62$\pm$0.20 & 0.58$\pm$0.26 & 0.63$\pm$0.83 \\
RADAb~\citep{wang2024radab} & 0.22$\pm$0.12 & 0.10$\pm$0.16 & -- & 11.57$\pm$49.30 & 0.10$\pm$0.09 & 0.48$\pm$0.33 & 7.91$\pm$34.02 & 0.57$\pm$0.23 & 0.57$\pm$0.28 & \underline{0.41$\pm$0.59} \\
dyAb~\citep{tan2025dyab} & 0.29$\pm$0.11 & 0.15$\pm$0.20 & -- & 2.89$\pm$0.88 & 0.07$\pm$0.03 & 0.51$\pm$0.29 & 1.92$\pm$0.82 & 0.64$\pm$0.16 & 0.65$\pm$0.28 & \textbf{0.00$\pm$0.00} \\
RefineGNN~\citep{jin2021refinegnn} & 0.23$\pm$0.14 & 0.13$\pm$0.18 & 7.91$\pm$3.20 & 2.85$\pm$0.88 & 0.09$\pm$0.05 & \textbf{0.65$\pm$0.27} & \textbf{1.38$\pm$0.77} & \textbf{0.73$\pm$0.16} & \textbf{0.78$\pm$0.20} & 0.64$\pm$0.79 \\
AbDockGen~\citep{jin2022hern} & 0.23$\pm$0.12 & 0.09$\pm$0.15 & 7.55$\pm$3.06 & 3.89$\pm$0.96 & 0.05$\pm$0.02 & 0.37$\pm$0.25 & 2.52$\pm$1.00 & 0.55$\pm$0.14 & 0.68$\pm$0.22 & 0.71$\pm$0.75 \\
AbODE~\citep{verma2023abode} & 0.37$\pm$0.13 & 0.19$\pm$0.22 & 9.26$\pm$4.09 & 15.63$\pm$4.55 & 0.00$\pm$0.00 & 0.07$\pm$0.12 & 4.65$\pm$2.22 & 0.35$\pm$0.10 & 0.19$\pm$0.22 & 0.83$\pm$0.39 \\
\bottomrule
\end{tabular}
\end{table}

\begin{table}[h!]
\centering
\caption{Generalization gap on CDR-H3. For each method, we report the epitope-group (EG, unseen epitope clusters) and antigen-fold (AF, unseen antigen folds) test splits. Sequence recovery (AAR) is nearly identical across the two splits, while the interface and epitope metrics (Fnat, DockQ, EpiF1) are consistently lower on the harder epitope-group split.}
\label{tab:gap}
\small
\setlength{\tabcolsep}{4pt}
\begin{tabular}{lcccccccc}
\toprule
 & \multicolumn{2}{c}{AAR$\uparrow$} & \multicolumn{2}{c}{Fnat$\uparrow$} & \multicolumn{2}{c}{DockQ$\uparrow$} & \multicolumn{2}{c}{EpiF1$\uparrow$} \\
\cmidrule(lr){2-3}\cmidrule(lr){4-5}\cmidrule(lr){6-7}\cmidrule(lr){8-9}
Method & EG & AF & EG & AF & EG & AF & EG & AF \\
\midrule
RAAD & 0.38 & 0.38 & 0.49 & 0.62 & 0.64 & 0.70 & 0.67 & 0.78 \\
MEAN & 0.42 & 0.40 & 0.48 & 0.62 & 0.63 & 0.70 & 0.67 & 0.78 \\
DiffAb & 0.22 & 0.23 & 0.48 & 0.52 & 0.58 & 0.58 & 0.56 & 0.65 \\
AbFlowNet & 0.22 & 0.23 & 0.49 & 0.56 & 0.57 & 0.60 & 0.55 & 0.65 \\
AbMEGD & 0.22 & 0.24 & 0.50 & 0.55 & 0.57 & 0.59 & 0.56 & 0.65 \\
RADAb & 0.22 & 0.25 & 0.47 & 0.51 & 0.57 & 0.57 & 0.57 & 0.65 \\
dyAb & 0.27 & 0.28 & 0.35 & 0.56 & 0.54 & 0.65 & 0.50 & 0.72 \\
RefineGNN & 0.23 & 0.26 & 0.62 & 0.60 & 0.71 & 0.68 & 0.76 & 0.80 \\
AbDockGen & 0.23 & 0.24 & 0.34 & 0.46 & 0.52 & 0.57 & 0.62 & 0.75 \\
AbODE & 0.31 & 0.27 & 0.08 & 0.16 & 0.34 & 0.39 & 0.20 & 0.40 \\
\bottomrule
\end{tabular}
\end{table}

\subsection{Cross-Method Metric Comparison}
\label{app:metric_comparison}

Table~\ref{tab:metric_cross} summarizes which metrics each of the surveyed methods originally reports, illustrating the inconsistency that \chimera{} standardizes. Abbreviations: DP = dynamic-programming RMSD (no Kabsch), K = Kabsch-aligned RMSD, sc = self-consistency RMSD (re-fold then compare), SeqID = BLOSUM62 alignment identity, PLL = pseudo-log-likelihood, LL/NLL = model (negative) log-likelihood.

\begin{table}[h!]
\centering
\caption{Metrics originally reported by each method (prior to \chimera{} standardization).}
\label{tab:metric_cross}
\small
\setlength{\tabcolsep}{3.5pt}
\begin{tabular}{lcccccccc}
\toprule
Method & AAR & RMSD & TM & lDDT & DockQ & PPL & dG/IMP & CAAR \\
\midrule
MEAN~\citep{kong2022mean}         & AAR   & K  & \checkmark & -- & -- & \checkmark & ddG & -- \\
dyMEAN~\citep{kong2023dymean}       & AAR   & K  & \checkmark & \checkmark & \checkmark & -- & ddG & \checkmark \\
RAAD~\citep{wu2025raad}         & AAR   & K  & \checkmark & -- & -- & \checkmark & -- & -- \\
RefineGNN~\citep{jin2021refinegnn}    & AAR   & K  & -- & -- & -- & \checkmark & -- & -- \\
AbODE~\citep{verma2023abode}        & AAR   & K  & -- & -- & -- & \checkmark & -- & -- \\
AbDockGen~\citep{jin2022hern}    & AAR   & K  & -- & -- & \checkmark & \checkmark & -- & -- \\
DiffAb~\citep{luo2022diffab}       & SeqID & DP & -- & -- & -- & -- & IMP,dG & -- \\
dyAb~\citep{tan2025dyab}         & AAR   & K  & \checkmark & \checkmark & \checkmark & -- & ddG & \checkmark \\
RADAb~\citep{wang2024radab}        & AAR & sc & -- & -- & -- & PLL & ddG & -- \\
AbFlowNet~\citep{abir2025abflownet}    & SeqID & DP & -- & -- & -- & -- & IMP,dG & -- \\
AbX~\citep{zhu2024abx}          & AAR   & K  & -- & -- & -- & PLL & IMP & -- \\
AbEgDiffuser~\citep{zhu2025abegdiffuser} & AAR   & DP & -- & -- & -- & -- & IMP & -- \\
AbMEGD~\citep{chen2025AbMEGD}       & SeqID & DP & -- & -- & -- & -- & IMP & -- \\
AbSGM~\citep{xie2023antibody-sgm}        & --    & K  & -- & -- & -- & -- & -- & -- \\
LEAD~\citep{yao2025lead} & AAR & DP & -- & -- & -- & -- & ddG & -- \\
ProteinMPNN~\citep{dauparas2022proteinmpnn} & AAR   & -- & -- & -- & -- & NLL & -- & -- \\
IgGM~\citep{wang2025iggm} & AAR   & K  & \checkmark & \checkmark & \checkmark & -- & -- & -- \\
DiffAbXL~\citep{uccar2024exploring} & --    & -- & -- & -- & -- & LL  & -- & -- \\
\bottomrule
\end{tabular}
\end{table}

\subsection{Baseline Retraining on \chimera{}}
\label{app:retraining}

All baselines are retrained on \chimera{} using the authors' released code with default hyperparameters. We provide a shared training and evaluation framework that handles data loading in each method's native input format, split-aware partitioning, early stopping, model checkpointing, and performs full evaluation. At test time, per-complex predictions are evaluated against native structures using all metrics defined in Section~\ref{sec:evaluation}. All training and inference is performed on an NVIDIA H100 GPU. Below we describe the method-specific adaptations required for each baseline.

\paragraph{DiffAb.}
DiffAb~\citep{luo2022diffab} operates under the Chothia numbering scheme and generates all six CDRs simultaneously by randomly masking between one and six CDRs during training. At inference, all CDRs are masked and regenerated in a single forward pass, producing one set of predictions per complex from which per-CDR metrics are extracted. We train with Adam at a learning rate of $10^{-4}$ and a batch size of 16, using 100 diffusion steps with equal loss weights on rotation, position, and sequence terms.

\paragraph{RAAD.}
RAAD~\citep{wu2025raad} generates one CDR at a time, requiring three training runs per split. We use the Adam optimizer with a learning rate of $10^{-3}$ and a batch size of 8. The model employs four relation-aware equivariant message-passing layers and constructs multi-relational graphs with radial, $k$-nearest neighbor, sequential, and global edge types. The structure loss weight is set to $\alpha{=}0.8$, placing greater emphasis on coordinate accuracy compared to MEAN.

\paragraph{MEAN.}
Because MEAN~\citep{kong2022mean} generates a single CDR loop per forward pass, we train three separate models per split to cover H1, H2, and H3. Each model is optimized with Adam at a learning rate of $10^{-3}$ and a batch size of 16. The structure loss is weighted by $\alpha{=}0.05$ relative to the sequence loss, and the model performs five multi-channel equivariant attention iterations per forward pass to progressively refine the generated CDR.

\paragraph{dyMEAN.}
dyMEAN~\citep{kong2023dymean} supports flexible CDR selection and performs three iterative refinement rounds per forward pass. The learning rate decays exponentially from $10^{-3}$ to $10^{-4}$ with a batch size of 4. A cosine annealing schedule linearly decreases the fraction of native template information provided to the model, gradually forcing reliance on its own predictions. The paratope is defined using a 6.6~\AA{} contact threshold under IMGT numbering.

\paragraph{RefineGNN.}
RefineGNN~\citep{jin2021refinegnn} generates CDR residues autoregressively from left to right while iteratively refining the predicted backbone structure after each residue is placed. Like MEAN and RAAD, it generates one CDR per forward pass, so we train three separate models per split for H1, H2, and H3. The model uses four message-passing layers with a hidden dimension of 256 and constructs $k$-nearest-neighbor graphs with $k{=}9$. A bidirectional GRU encodes framework context, and the structural refinement step runs at every position (update frequency 1), predicting pairwise distances that are converted to coordinates via eigendecomposition. During training, teacher forcing provides the ground-truth sequence at each step, while at inference, the model samples from the predicted distribution autoregressively.

\paragraph{AbDockGen.}
AbDockGen~\citep{jin2022hern} is a hierarchical equivariant refinement network that generates CDR-H3 conditioned on the epitope surface, making it applicable only to \textbf{Track~1} evaluation. The model encodes the top 20 nearest epitope residues using an E(3)-equivariant graph neural network and generates the CDR-H3 sequence and structure autoregressively with iterative coordinate refinement. The architecture consists of four hierarchical EGNN layers with a hidden dimension of 256 and $k{=}9$ nearest neighbors, operating at both atom and residue levels. Explicit clash avoidance constraints enforce minimum distances of 3.8~\AA{} between backbone atoms and 1.5~\AA{} between side-chain atoms during coordinate updates. We train with Adam at a learning rate of $10^{-3}$, dynamic batching at 100 tokens, and gradient clipping at norm 1.0, for at most 10 epochs with a learning rate anneal factor of 0.9. 

\paragraph{AbODE.}
AbODE~\citep{verma2023abode} formulates CDR generation as a conjoined ordinary differential equation that jointly evolves sequence logits and backbone coordinates. Like MEAN and RAAD, it generates one CDR at a time and requires three training runs per split. The model consists of four TransformerConv layers and we train with Adam at a learning rate of $10^{-3}$ and a batch size of 1 due to variable-size fully-connected graphs.

\paragraph{AbFlowNet.}
AbFlowNet~\citep{abir2025abflownet} extends DiffAb's diffusion architecture with a GFlowNet trajectory balance objective to optimize binding energy. It uses Chothia numbering and generates all six CDRs simultaneously. The training-time backward loss is disabled for CHIMERA integration. We train with Adam at a learning rate of $10^{-4}$ and a batch size of 16.

\paragraph{AbMEGD.}
AbMEGD~\citep{chen2025AbMEGD} is a multi-CDR diffusion model that shares DiffAb's Chothia-based preprocessing pipeline and generates all six CDRs simultaneously. We train with the authors' default hyperparameters using Adam optimization.

\paragraph{RADAb.}
RADAb~\citep{wang2024radab} is a retrieval-augmented diffusion model that combines DiffAb's diffusion framework with sequence retrieval from a FASTA database. It uses ESM-2 (650M parameters) for sequence embeddings and an MSA transformer for retrieved sequence alignment. We train under Chothia numbering with Adam optimization.

\paragraph{dyAb.}
dyAb~\citep{tan2025dyab} applies flow matching to flexible antibody design under IMGT numbering, generating all six CDRs simultaneously. The model includes OpenMM-based structure relaxation and a DDG prediction module. We train with the authors' default configuration.

\subsection{Protein Graph Construction}
\label{app:graphs}

For each complex, we construct a heterogeneous multi-relational residue graph~\citep{wu2025raad} stored as a PyG \texttt{HeteroData} object. The graph contains three node types corresponding to heavy chain, light chain, and antigen residues.

\paragraph{Node features (105D).} Each residue node carries a 105-dimensional feature vector consisting of a residue type one-hot encoding (20D), sinusoidal positional encoding along the chain (16D), backbone dihedral angle sin/cos encodings (12D for 6 angles), radial basis function (RBF) encodings of C$\alpha$-to-backbone-atom distances (48D for 3 atoms with 16 RBF bins each), and a local coordinate frame derived from N, C$\alpha$, C atoms via Gram--Schmidt orthogonalization (9D). Each node additionally stores the residue type index, backbone atom coordinates (N, C$\alpha$, C, O), CDR region labels under IMGT numbering, a binary flag marking designable CDR positions, and binary epitope or paratope labels. Since pre-trained protein language model (PLM) embeddings are proven to capture meaningful evolutionary features of proteins~\citep{ahmed2025improved}, we utilize ESM2~\citep{lin2023evolutionary} to compute embeddings for antigens and AntiBERTy~\citep{ruffolo2023igfold} for antibodies.

\paragraph{Intra-chain edges.} Within each chain, four relation types encode complementary structural relationships. Sequential $\pm$1 edges connect adjacent residues along the backbone, sequential $\pm$2 edges provide broader backbone context, $k$-nearest neighbor edges ($k{=}10$) capture local 3D proximity regardless of sequence position, and spatial edges connect all residue pairs within 8.0~\AA{} not already covered by the preceding types. Each intra-chain edge carries a 39-dimensional feature vector consisting of edge type one-hot (4D), relative positional encoding (16D), C$\alpha$--C$\alpha$ distance RBF encoding (16D), and normalized direction vector (3D).

\paragraph{Inter-chain edges.} Two types of inter-chain edges connect antibody and antigen nodes. Contact edges are derived from the annotated contact pairs (4.5~\AA{} heavy-atom cutoff) and directly encode the binding interface by linking heavy-to-antigen and light-to-antigen residue pairs. Spatial edges use a broader 12~\AA{} C$\alpha$--C$\alpha$ cutoff with 16-bin RBF distance features, providing the spatial context used by MEAN, dyMEAN, dyAb, and RAAD for message passing across chains. Heavy-to-light spatial edges are also included for intra-antibody context.

The resulting graphs are saved as a single dictionary mapping complex identifiers to \texttt{HeteroData} objects. These precomputed graphs can be used directly by graph-based methods or converted to homogeneous representations as needed.

\end{document}

%% file: math_commands.tex
\usepackage{amsmath,amsfonts,bm}

\def\eqref#1{equation~\ref{#1}}

\def\1{\bm{1}}

\DeclareMathAlphabet{\mathsfit}{\encodingdefault}{\sfdefault}{m}{sl}
\SetMathAlphabet{\mathsfit}{bold}{\encodingdefault}{\sfdefault}{bx}{n}

\DeclareMathOperator*{\argmax}{arg\,max}

%% file: references.bib
@article{ahmed2025improved,
	author = {Ahmed, Mansoor and Ali, Sarwan and Jan, Avais and Khan, Imdad Ullah and Patterson, Murray},
	title = {Improved Graph-based Antibody-aware Epitope Prediction with Protein Language Model-based Embeddings},
	elocation-id = {2025.02.12.637989},
	year = {2025},
	doi = {10.1101/2025.02.12.637989},
	publisher = {Cold Spring Harbor Laboratory},
	journal = {bioRxiv}
}

@article{xue2015computational,
  title={Computational prediction of protein interfaces: A review of data driven methods},
  author={Xue, Li C and Dobbs, Drena and Bonvin, Alexandre MJJ and Honavar, Vasant},
  journal={FEBS letters},
  volume={589},
  number={23},
  pages={3516--3526},
  year={2015},
  publisher={Elsevier}
}

@article{janin2003capri,
  title={{CAPRI}: a critical assessment of predicted interactions},
  author={Janin, Jo{\"e}l and Henrick, Kim and Moult, John and Eyck, Lynn Ten and Sternberg, Michael JE and Vajda, Sandor and Vakser, Ilya and Wodak, Shoshana J},
  journal={Proteins: Structure, Function, and Bioinformatics},
  volume={52},
  number={1},
  pages={2--9},
  year={2003},
  publisher={Wiley Online Library}
}

@article{ovchinnikov2014fnat,
  title={Robust and accurate prediction of residue--residue interactions across protein interfaces using evolutionary information},
  author={Ovchinnikov, Sergey and Kamisetty, Hetunandan and Baker, David},
  journal={elife},
  volume={3},
  pages={e02030},
  year={2014},
  publisher={eLife Sciences Publications, Ltd}
}

@article{zhao2025abbibench,
  title={Ab{B}i{B}ench: A Benchmark for Antibody Binding Affinity Maturation and Design},
  author={Zhao, Xinyan and Tang, Yi-Ching and Singh, Akshita and Cantu, Victor J and An, KwanHo and Lee, Junseok and Stogsdill, Adam E and Hamdi, Ibraheem M and Ramesh, Ashwin Kumar and An, Zhiqiang and others},
  journal={arXiv preprint arXiv:2506.04235},
  year={2025}
}

@article{liu2024asep,
  title={{AsEP}: Benchmarking Deep Learning Methods for Antibody-specific Epitope Prediction},
  author={Liu, Chunan and Denzler, Lilian and Chen, Yihong and Martin, Andrew and Paige, Brooks},
  journal={arXiv preprint arXiv:2407.18184},
  year={2024}
}

@article{ruffolo2023igfold,
  title={Fast, accurate antibody structure prediction from deep learning on massive set of natural antibodies},
  author={Ruffolo, Jeffrey A and Chu, Lee-Shin and Mahajan, Sai Pooja and Gray, Jeffrey J},
  journal={Nature communications},
  volume={14},
  number={1},
  pages={2389},
  year={2023},
  publisher={Nature Publishing Group UK London}
}

@article{norman2020computational,
  title={Computational approaches to therapeutic antibody design: established methods and emerging trends},
  author={Norman, Richard and Ambrosetti, Francesco and Bonvin, Alexandre and Colwell, Lucy and Kelm, Sebastian and Kumar, Sandeep and Krawczyk, Konrad},
  journal={Briefings in bioinformatics},
  volume={21},
  number={5},
  pages={1549--1567},
  year={2020},
  publisher={Oxford University Press}
}

@inproceedings{zhu2024abx,
  title={Antibody design using a score-based diffusion model guided by evolutionary, physical and geometric constraints},
  author={Zhu, Tian and Ren, Milong and Zhang, Haicang},
  booktitle={Forty-first International Conference on Machine Learning},
  year={2024}
}

@article{yao2025lead,
  title={Generative co-design of antibody sequences and structures via black-box guidance in a shared latent space},
  author={Yao, Yinghua and Pan, Yuangang and Chen, Xixian},
  journal={arXiv preprint arXiv:2508.11424},
  year={2025}
}

@article{wang2024radab,
  title={Retrieval augmented diffusion model for structure-informed antibody design and optimization},
  author={Wang, Zichen and Ji, Yaokun and Tian, Jianing and Zheng, Shuangjia},
  journal={arXiv preprint arXiv:2410.15040},
  year={2024}
}

@inproceedings{verma2023abode,
  title={Ab{ODE}: Ab {I}nitio antibody design using conjoined {ODE}s},
  author={Verma, Yogesh and Heinonen, Markus and Garg, Vikas},
  booktitle={International Conference on Machine Learning},
  pages={35037--35050},
  year={2023},
  organization={PMLR}
}

@article{jin2021refinegnn,
  title={Iterative refinement graph neural network for antibody sequence-structure co-design},
  author={Jin, Wengong and Wohlwend, Jeremy and Barzilay, Regina and Jaakkola, Tommi},
  journal={arXiv preprint arXiv:2110.04624},
  year={2021}
}

@inproceedings{wu2025raad,
  title={Relation-aware equivariant graph networks for epitope-unknown antibody design and specificity optimization},
  author={Wu, Lirong and Lin, Haitao and Huang, Yufei and Gao, Zhangyang and Tan, Cheng and Liu, Yunfan and Wu, Tailin and Li, Stan Z},
  booktitle={Proceedings of the AAAI Conference on Artificial Intelligence},
  volume={39},
  number={1},
  pages={895--904},
  year={2025}
}

@inproceedings{jin2022hern,
  title={Antibody-antigen docking and design via hierarchical structure refinement},
  author={Jin, Wengong and Barzilay, Regina and Jaakkola, Tommi},
  booktitle={International Conference on Machine Learning},
  pages={10217--10227},
  year={2022},
  organization={PMLR}
}

@article{bennett2025rfdiffusion-ab,
  title={Atomically accurate de novo design of antibodies with {RF}diffusion},
  author={Bennett, Nathaniel R and Watson, Joseph L and Ragotte, Robert J and Borst, Andrew J and See, D{\'e}Jena{\'e} L and Weidle, Connor and Biswas, Riti and Yu, Yutong and Shrock, Ellen L and Ault, Russell and others},
  journal={Nature},
  pages={1--11},
  year={2025},
  publisher={Nature Publishing Group UK London}
}

@article{lin2023evolutionary,
  title={Evolutionary-scale prediction of atomic-level protein structure with a language model},
  author={Lin, Zeming and Akin, Halil and Rao, Roshan and Hie, Brian and Zhu, Zhongkai and Lu, Wenting and Smetanin, Nikita and Verkuil, Robert and Kabeli, Ori and Shmueli, Yaniv and others},
  journal={Science},
  volume={379},
  number={6637},
  pages={1123--1130},
  year={2023},
  publisher={American Association for the Advancement of Science}
}

@article{kong2023dymean,
  title={End-to-end full-atom antibody design},
  author={Kong, Xiangzhe and Huang, Wenbing and Liu, Yang},
  journal={arXiv preprint arXiv:2302.00203},
  year={2023}
}

@article{luo2022diffab,
  title={Antigen-specific antibody design and optimization with diffusion-based generative models for protein structures},
  author={Luo, Shitong and Su, Yufeng and Peng, Xingang and Wang, Sheng and Peng, Jian and Ma, Jianzhu},
  journal={Advances in Neural Information Processing Systems},
  volume={35},
  pages={9754--9767},
  year={2022}
}

@inproceedings{tan2025dyab,
  title={dy{A}b: Flow Matching for Flexible Antibody Design with {A}lpha{F}old-driven Pre-binding Antigen},
  author={Tan, Cheng and Zhang, Yijie and Gao, Zhangyang and Huang, Yufei and Lin, Haitao and Wu, Lirong and Wu, Fandi and Blanchette, Mathieu and Li, Stan Z},
  booktitle={Proceedings of the AAAI Conference on Artificial Intelligence},
  volume={39},
  number={1},
  pages={782--790},
  year={2025}
}

@inproceedings{uccar2024exploring,
  title={Exploring log-likelihood scores for ranking antibody sequence designs},
  author={U{\c{c}}ar, Talip and Malherbe, Cedric and Gonzalez, Ferran},
  booktitle={NeurIPS 2024 Workshop on AI for New Drug Modalities},
  year={2024}
}

@article{kong2022mean,
  title={Conditional antibody design as 3{D} equivariant graph translation},
  author={Kong, Xiangzhe and Huang, Wenbing and Liu, Yang},
  journal={arXiv preprint arXiv:2208.06073},
  year={2022}
}

@article{wang2025iggm,
  title={A generative foundation model for antibody design},
  author={Wang, Rubo and Wu, Fandi and Shi, Jiale and Song, Yidong and Kong, Yu and Ma, Jian and He, Bing and Yan, Qihong and Ying, Tianlei and Zhao, Peilin and others},
  journal={bioRxiv},
  pages={2025--09},
  year={2025},
  publisher={Cold Spring Harbor Laboratory}
}

@article{abir2025abflownet,
  title={Ab{F}low{N}et: Optimizing Antibody-Antigen Binding Energy via Diffusion-{GF}low{N}et Fusion},
  author={Abir, Abrar Rahman and Shahgir, Haz Sameen and Ratul, Md Rownok Zahan and Tahmid, Md Toki and Steeg, Greg Ver and Dong, Yue},
  journal={arXiv preprint arXiv:2505.12358},
  year={2025}
}

@article{chen2025AbMEGD,
  title={Antibody Design and Optimization with Multi-scale Equivariant Graph Diffusion Models for Accurate Complex Antigen Binding},
  author={Chen, Jiameng and Cai, Xiantao and Wu, Jia and Hu, Wenbin},
  journal={arXiv preprint arXiv:2506.20957},
  year={2025}
}

@article{martinkus2023abdiffuser,
  title={Ab{D}iffuser: full-atom generation of in-vitro functioning antibodies},
  author={Martinkus, Karolis and Ludwiczak, Jan and Liang, Wei-Ching and Lafrance-Vanasse, Julien and Hotzel, Isidro and Rajpal, Arvind and Wu, Yan and Cho, Kyunghyun and Bonneau, Richard and Gligorijevic, Vladimir and others},
  journal={Advances in Neural Information Processing Systems},
  volume={36},
  pages={40729--40759},
  year={2023}
}

@article{dauparas2022proteinmpnn,
  title={Robust deep learning--based protein sequence design using {P}rotein{MPNN}},
  author={Dauparas, Justas and Anishchenko, Ivan and Bennett, Nathaniel and Bai, Hua and Ragotte, Robert J and Milles, Lukas F and Wicky, Basile IM and Courbet, Alexis and de Haas, Rob J and Bethel, Neville and others},
  journal={Science},
  volume={378},
  number={6615},
  pages={49--56},
  year={2022},
  publisher={American Association for the Advancement of Science}
}

@inproceedings{xie2023antibody-sgm,
  title={Antibody-{SGM}: Antigen-Specific Joint Design of Antibody Sequence and Structure using Diffusion Models},
  author={Xie, Xuezhi and Lee, Jin Sub and Kim, Dongki and Jo, Jaehyeong and Kim, Jisun and Kim, Philip M},
  booktitle={2023 ICML Workshop Comput Biol},
  year={2023}
}

@article{dunbar2014sabdab,
  title={{SA}b{D}ab: the structural antibody database},
  author={Dunbar, James and Krawczyk, Konrad and Leem, Jinwoo and Baker, Terry and Fuchs, Angelika and Georges, Guy and Shi, Jiye and Deane, Charlotte M},
  journal={Nucleic acids research},
  volume={42},
  number={D1},
  pages={D1140--D1146},
  year={2014},
  publisher={Oxford University Press}
}

@article{adolf2018rabd,
  title={Rosetta {A}ntibody {D}esign ({RA}b{D}): A general framework for computational antibody design},
  author={Adolf-Bryfogle, Jared and Kalyuzhniy, Oleks and Kubitz, Michael and Weitzner, Brian D and Hu, Xiaozhen and Adachi, Yumiko and Schief, William R and Dunbrack Jr, Roland L},
  journal={PLoS computational biology},
  volume={14},
  number={4},
  pages={e1006112},
  year={2018},
  publisher={Public Library of Science San Francisco, CA USA}
}

@article{notin2023proteingym,
  title={Protein{G}ym: Large-scale benchmarks for protein fitness prediction and design},
  author={Notin, Pascal and Kollasch, Aaron and Ritter, Daniel and Van Niekerk, Lood and Paul, Steffanie and Spinner, Han and Rollins, Nathan and Shaw, Ada and Orenbuch, Rose and Weitzman, Ruben and others},
  journal={Advances in Neural Information Processing Systems},
  volume={36},
  pages={64331--64379},
  year={2023}
}

@article{zhang2004tm,
  title={Scoring function for automated assessment of protein structure template quality},
  author={Zhang, Yang and Skolnick, Jeffrey},
  journal={Proteins: Structure, Function, and Bioinformatics},
  volume={57},
  number={4},
  pages={702--710},
  year={2004},
  publisher={Wiley Online Library}
}

@article{basu2016dockq,
  title={Dock{Q}: a quality measure for protein-protein docking models},
  author={Basu, Sankar and Wallner, Bj{\"o}rn},
  journal={PloS one},
  volume={11},
  number={8},
  pages={e0161879},
  year={2016},
  publisher={Public Library of Science San Francisco, CA USA}
}

@article{steinegger2017mmseqs2,
  title={{MM}seqs2 enables sensitive protein sequence searching for the analysis of massive data sets},
  author={Steinegger, Martin and S{\"o}ding, Johannes},
  journal={Nature biotechnology},
  volume={35},
  number={11},
  pages={1026--1028},
  year={2017},
  publisher={Nature Publishing Group US New York}
}

@article{dunbar2016anarci,
  title={{ANARCI}: antigen receptor numbering and receptor classification},
  author={Dunbar, James and Deane, Charlotte M},
  journal={Bioinformatics},
  volume={32},
  number={2},
  pages={298--300},
  year={2016},
  publisher={Oxford University Press}
}

@article{zhu2025abegdiffuser,
  title={Ab{E}g{D}iffuser: Antibody Sequence-Structure Codesign with Equivariant Graph Neural Networks and Diffusion Models},
  author={Zhu, Yibo and Shi, Xiumin and Zhang, Jingjuan and Sun, Weizhong and Wang, Lu},
  journal={Journal of Chemical Theory and Computation},
  volume={21},
  number={21},
  pages={11307--11317},
  year={2025},
  publisher={ACS Publications}
}

@article{chothia1987canonical,
  title={Canonical structures for the hypervariable regions of immunoglobulins},
  author={Chothia, Cyrus and Lesk, Arthur M},
  journal={Journal of molecular biology},
  volume={196},
  number={4},
  pages={901--917},
  year={1987},
  publisher={Elsevier}
}

@article{lefranc2003imgt,
  title={{IMGT} unique numbering for immunoglobulin and {T} cell receptor variable domains and {Ig} superfamily {V}-like domains},
  author={Lefranc, Marie-Paule and Pommi{\'e}, Christelle and Ruiz, Manuel and Giudicelli, V{\'e}ronique and Foulquier, Elodie and Truong, Lisa and Thouvenin-Contet, Val{\'e}rie and Lefranc, G{\'e}rard},
  journal={Developmental \& Comparative Immunology},
  volume={27},
  number={1},
  pages={55--77},
  year={2003},
  publisher={Elsevier}
}

@article{townshend2020atom3d,
  title={{ATOM3D}: Tasks on molecules in three dimensions},
  author={Townshend, Raphael JL and V{\"o}gele, Martin and Suriana, Patricia and Derry, Alexander and Powers, Alexander and Laloudakis, Yianni and Balachandar, Sidhika and Jing, Bowen and Anderson, Brandon and Eismann, Stephan and others},
  journal={arXiv preprint arXiv:2012.04035},
  year={2020}
}

@article{jankauskaite2019skempi,
  title={{SKEMPI} 2.0: an updated benchmark of changes in protein--protein binding energy, kinetics and thermodynamics upon mutation},
  author={Jankauskait{\.e}, Justina and Jim{\'e}nez-Garc{\'\i}a, Brian and Dapk{\=u}nas, Justas and Fern{\'a}ndez-Recio, Juan and Moal, Iain H},
  journal={Bioinformatics},
  volume={35},
  number={3},
  pages={462--469},
  year={2019},
  publisher={Oxford University Press}
}

@article{stark2025boltzgen,
  title={Boltz{G}en: Toward universal binder design},
  author={Stark, Hannes and Faltings, Felix and Choi, MinGyu and Xie, Yuxin and Hur, Eunsu and O’Donnell, Timothy and Bushuiev, Anton and U{\c{c}}ar, Talip and Passaro, Saro and Mao, Weian and others},
  journal={bioRxiv},
  pages={2025--11},
  year={2025},
  publisher={Cold Spring Harbor Laboratory}
}

@article{ramaraj2012antigen,
  title={Antigen--antibody interface properties: Composition, residue interactions, and features of the antigen},
  author={Ramaraj, Thiru and Angel, Thomas and Dratz, Edward A and Jesaitis, Algirdas J and Bhattacharyya, Suvobrata},
  journal={Biochimica et Biophysica Acta (BBA) -- Proteins and Proteomics},
  volume={1824},
  number={3},
  pages={520--532},
  year={2012},
  publisher={Elsevier}

}

@incollection{boerner2023access,
  title={Access: Advancing innovation: {NSF}’s advanced cyberinfrastructure coordination ecosystem: Services \& support},
  author={Boerner, Timothy J and Deems, Stephen and Furlani, Thomas R and Knuth, Shelley L and Towns, John},
  booktitle={Practice and experience in advanced research computing 2023: Computing for the common good},
  pages={173--176},
  year={2023}
}
